\documentclass[lettersize,journal]{IEEEtran}
\usepackage{amsmath,amsfonts}

\usepackage{algorithm}
\usepackage{algpseudocode}
\usepackage{array}
\usepackage[caption=false,font=normalsize,labelfont=sf,textfont=sf]{subfig}
\usepackage{textcomp}
\usepackage{stfloats}
\usepackage{url}
\usepackage{verbatim}
\usepackage{graphicx}
\usepackage{cite}
\usepackage{tabularx}
\usepackage{booktabs}
\usepackage{multirow} 
\begin{document}

\title{Utilizing Multiple Inputs Autoregressive Models for Bearing Remaining Useful Life Prediction}

\author{Junliang Wang $^{1,2}$, Qinghua Zhang $^{3,}$* , Guanhua Zhu$^{3,}$*  and Guoxi Sun$^{3}$}



\maketitle

\begin{abstract}
Accurate prediction of the Remaining Useful Life (RUL) of rolling bearings is crucial in industrial production, yet existing models often struggle with limited generalization capabilities due to their inability to fully process all vibration signal patterns. We introduce a novel multi-input autoregressive model to address this challenge in RUL prediction for bearings. Our approach uniquely integrates vibration signals with previously predicted Health Indicator (HI) values, employing feature fusion to output current window HI values. Through autoregressive iterations, the model attains a global receptive field, effectively overcoming the limitations in generalization. Furthermore, we innovatively incorporate a segmentation method and multiple training iterations to mitigate error accumulation in autoregressive models. Empirical evaluation on the PMH2012 dataset demonstrates that our model, compared to other backbone networks using similar autoregressive approaches, achieves significantly lower Root Mean Square Error (RMSE) and Score. Notably, it outperforms traditional autoregressive models that use label values as inputs and non-autoregressive networks, showing superior generalization abilities with a marked lead in RMSE and Score metrics.
\end{abstract}


\section{Introduction}
\IEEEPARstart{W}{ith} the rapid development of modern industry, Prognostics and Health Management (PHM) of machinery plays an irreplaceable role in enhancing the reliability and efficiency of smart factory operations. Bearings, as an essential component of modern production machinery, have high demands for reliability and safety and are one of the primary causes of mechanical failures \cite{re1}. Therefore, predicting and preventing failures by estimating the Remaining Useful Life (RUL) of bearings is crucial. The degradation of machine tool gearbox bearings is related to multiple factors such as workload, operating modes, material defects, and operating temperature. These factors comprehensively affect the health status of bearings, making RUL prediction a challenging task \cite{re2}.
Typically, methods for predicting bearing RUL include model-based \cite{ref5,ref6,physical1,physical2}, data-driven, and hybrid approaches \cite{re3,hybrid1,hybrid2,hybrid3}. Compared to model-based methods, data-driven approaches can learn degradation patterns from sensor signals and establish predictive models through machine learning or deep learning. Moreover, with the rapid advancement of sensing technologies, acquiring a large volume of industrial data, such as vibration signals, has become easier. When the data volume is substantial, data-driven methods significantly outperform the other two approaches.

The prediction of Remaining Useful Life (RUL) is fundamentally a regression problem related to time series. The effectiveness of a designed model in learning key features of time series is crucial to the accuracy of the final prediction\cite{re4}. To fully exploit temporal information, deep learning-based networks, such as Recurrent Neural Networks (RNN), Long Short-Term Memory networks (LSTM), and their variants, have been extensively applied in the field of Prognostics and Health Management (PHM)\cite{re5,ref22,ref7,ref17,ref18}. These networks are valued for their proficiency in capturing the dynamic characteristics and dependencies in time-series data, which is essential for precise RUL estimation.

Guo et al. \cite{ref22} investigated the use of Recurrent Neural Networks (RNN) for deriving health indicators in bearing Remaining Useful Life (RUL) prediction. Zheng et al. \cite{zo1} proposed a method for RUL prediction using Long Short-Term Memory (LSTM) networks. The authors demonstrated the effectiveness of the LSTM model in analyzing sensor data from aircraft engines, particularly in handling long sequence data.
Chen et al. \cite{zo2} employed Gate Recurrent Unit (GRU) models in place of LSTM for RUL prediction during nonlinear degradation processes. They reported improved performance in terms of prediction accuracy and training time.
These studies indicate that deep learning, particularly Recurrent Neural Networks (RNN) and Long Short-Term Memory (LSTM) networks, have significant advantages in tasks like Remaining Useful Life (RUL) prediction and other time-series analyses.

Convolutional Neural Networks (CNNs) have found extensive applications in deep learning. Currently, there are two main approaches for utilizing CNNs in bearing remaining life prediction. The first method involves employing one-dimensional convolution, directly processing vibration signals in the time domain. The second method transforms vibration signals into two-dimensional time-frequency spectrograms and then applies two-dimensional convolution. Both approaches yield promising results when implemented with appropriate network architectures.
Ren et al. \cite{ref16} proposed a novel approach for predicting the Remaining Useful Life (RUL) of bearings, utilizing Deep Convolutional Neural Networks (CNN). Li et al. \cite{zo3} introduced a Deep Convolutional Neural Network (DCNN) using normalized raw data as the network input. They validated the effectiveness of their method for RUL prediction in aircraft engine bearings.
In a separate study, Li et al. \cite{zo4} enhanced the network's capability to learn complex features by extracting features at different scales. They employed a Multi-Scale Deep Convolutional Neural Network (MSDCNN) for RUL prediction in engines.

The Temporal Convolutional Network (TCN) is a neural network architecture specifically designed for processing time-series data
\cite{tcn}. It represents a recent enhancement to the CNN framework, utilizing Dilated Causal Convolution (DCC) to extract information from historical data. 
Additionally, it leverages the strengths of sequential networks such as Recurrent Neural Networks (RNNs) and Long Short-Term Memory (LSTM) networks in modeling temporal dependencies, crucial for tasks involving time series data like language modeling or remaining useful life (RUL) prediction. Notably, the TCN model successfully addresses the issue of gradient explosion inherent in RNNs and LSTMs\cite{explode1,explode2}, ensuring stable and convergent training.This enhances the model's trainability and generalization performance.

Yi’an et al. \cite{zo5} proposed a bearing Remaining Useful Life (RUL) prediction model based on the TCN-Transformer. This model leverages the strengths of TCN and the capabilities of the Transformer model. Researchers have improved the loss function on the basis of TCN, introducing T-MSE, and established WTCN to predict the RUL of bearings. This method adequately accounts for the variations and severity in bearing vibration signals, thereby enhancing prediction accuracy \cite{zo6}.
In another study, a new framework called the Temporal Convolutional Network with residual separable convolutional block (TCN-RSA) was introduced for bearing RUL prediction. This framework employs residual learning and separable convolution to construct an effective convolutional block, aiming to improve predictive precision \cite{zo7}.

Existing RUL predictive methods fundamentally involve regression operations on vibration waveforms, which are challenging to apply in practical forecasting. Autoregressive methods are a tool for modeling time-series data, utilized to describe the relationship between a variable and its own past observed values. This characteristic makes autoregressive methods well-suited for fulfilling the requirements of predictive tasks in time-series data modeling. Such methods are common in time-series models like recurrent neural networks (RNNs) and long short-term memory networks (LSTMs), where they model sequences of information within a single sample.

In the field of natural language processing (NLP), the Transformer \cite{transformer} has achieved remarkable results using autoregressive models. The research by Liu et al. \cite{zo8} represents the latest advancement in autoregressive structured prediction. Their method has set new benchmarks in all observed structured prediction tasks, including named entity recognition, end-to-end relation extraction, and coreference resolution.

However, in the field of Prognostics and Health Management (PHM) of machinery, the application of autoregressive methods for degradation prediction remains relatively less explored. The focus within PHM has predominantly been on fault diagnosis tasks, and these are typically anchored in physical modeling approaches.

Al-Bugharbee's approach\cite{ar1} relies on linear time-invariant autoregressive modeling for rolling element bearings fault diagnosis. It emphasizes the importance of signal pretreatment, incorporating noise reduction through singular spectrum analysis and stationarization via differencing to enhance diagnostic accuracy.
Ma et al.\cite{ar2} introduces a novel hybrid model (GNAR-GARCH) for rolling bearing fault diagnosis, combining linear and nonlinear autoregressive features with a generalized autoregressive conditional heteroscedasticity model. This model excels in handling nonlinear and nonstationary signals, demonstrating superior accuracy and performance compared to other models.
Qian et al.\cite{ar3} presents a multi-time scale approach integrating enhanced phase space warping (PSW) and a modified Paris crack growth model for bearing defect tracking and remaining useful life (RUL) prediction. This data-driven method combines PSW for fast-time scale behavior and a physics-based crack growth model for slow-time scale characterization.
Nistane et al.\cite{ar4} proposes a health assessment model for rotary machine rolling element bearings based on a non-linear autoregressive neural network and the exponential value of health indicator (HI). Using vibration signals and continuous wavelet transform for feature extraction, the model achieves accurate degradation prediction through optimal NAR and NARX networks, demonstrating effectiveness in various scenarios.

\begin{figure*}[!t]
\centering
\subfloat[]{\includegraphics[width=2in]{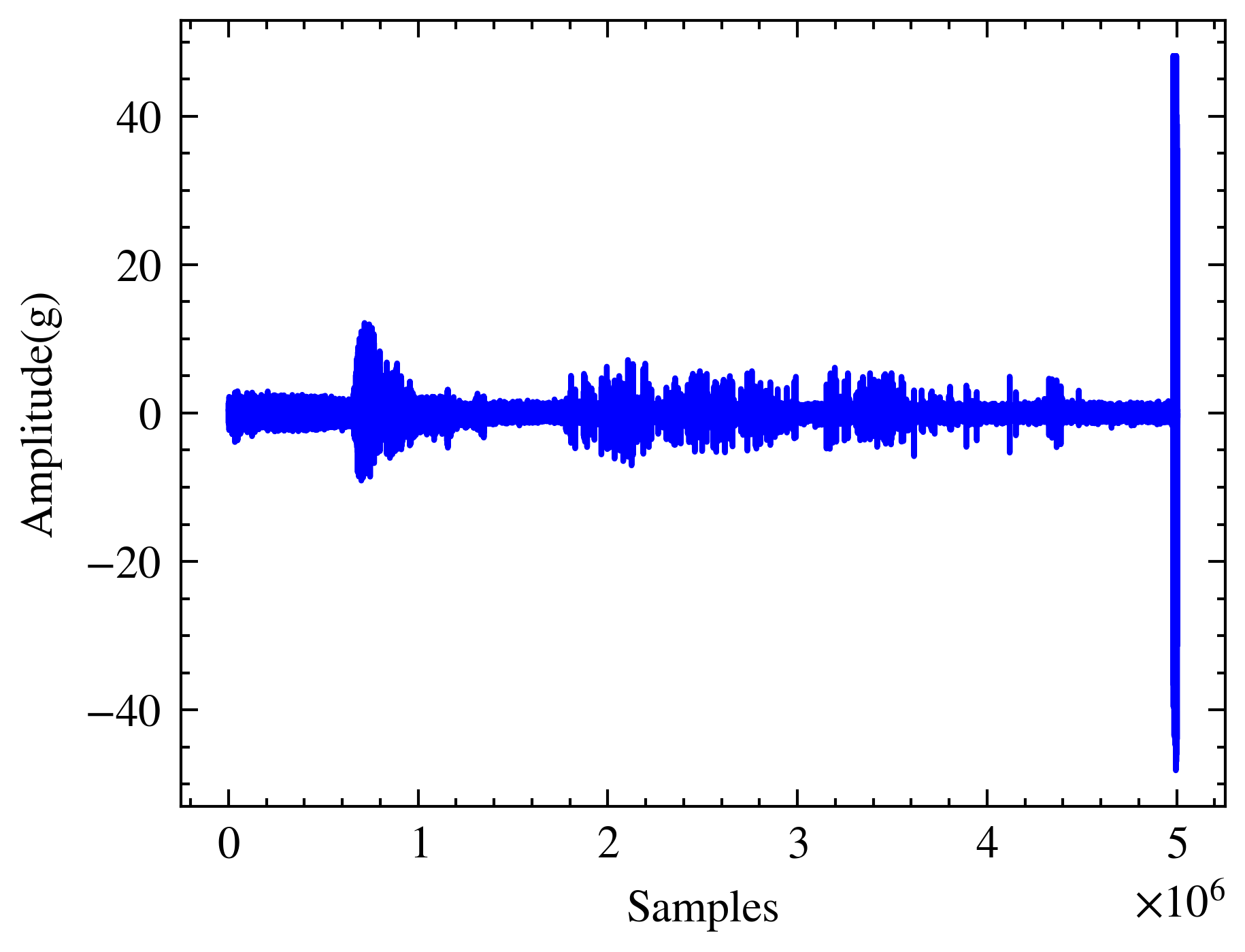}%
\label{fig:1a}}
\hfil
\subfloat[]{\includegraphics[width=2in]{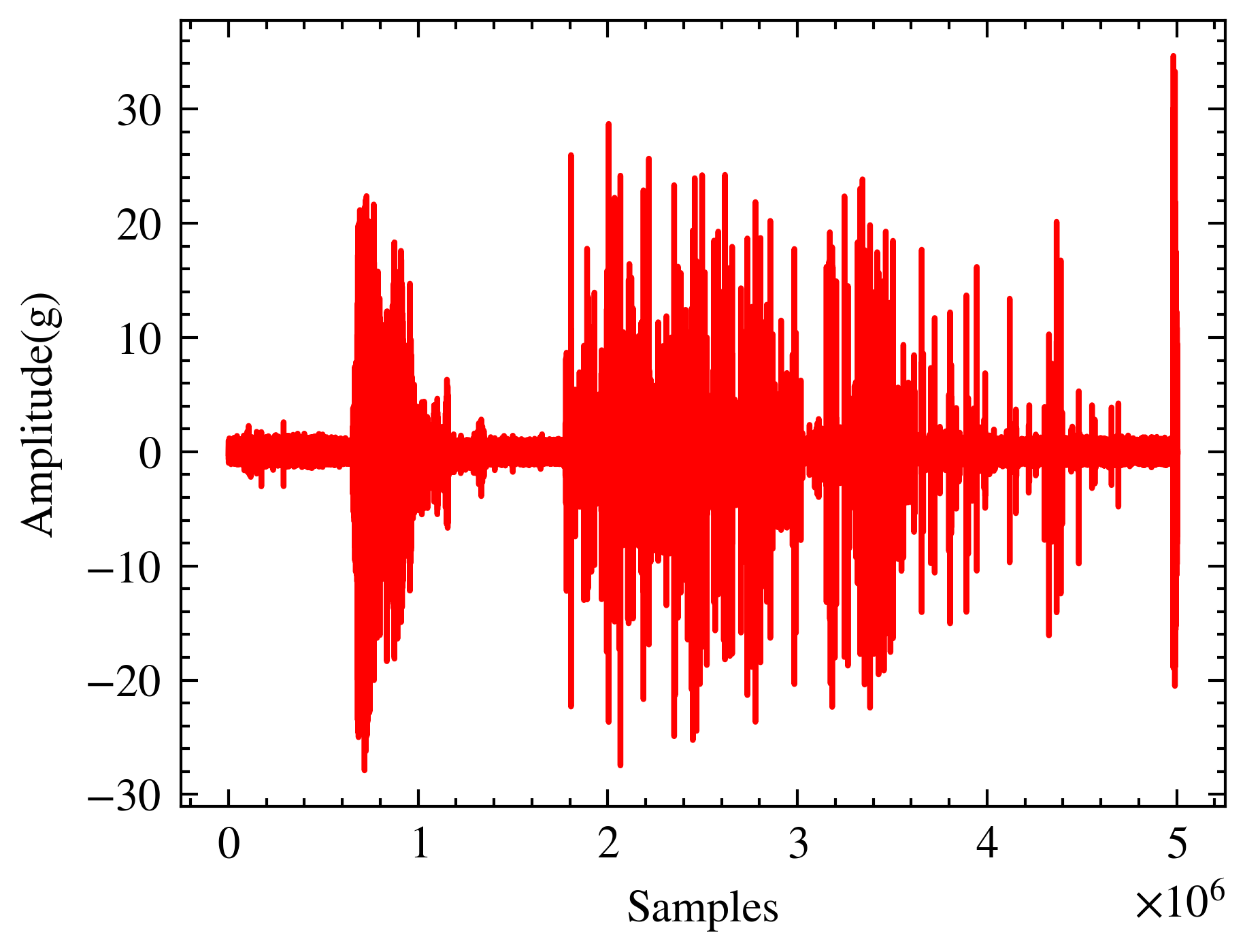}%
\label{fig:1b}}
\hfil
\subfloat[]{\includegraphics[width=2in]{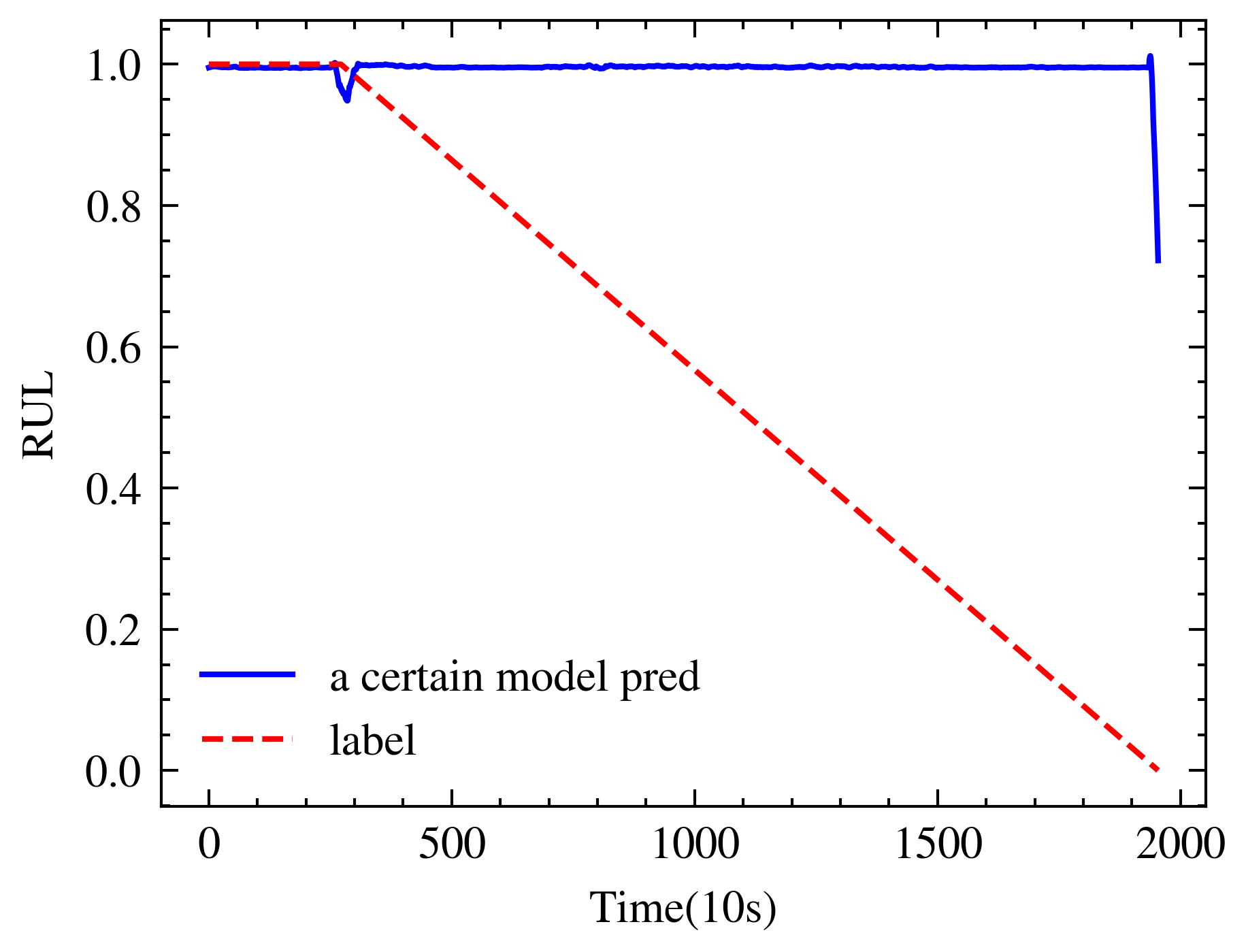}%
\label{fig:1c}}
\caption{An example of RUL prediction failure.  }
\label{fig_sim}
\end{figure*}
Based on the aforementioned analysis, to the best of our knowledge, bearing life prediction utilizing autoregressive models is still in its infancy. In this paper, we also want to share a common scenario often encountered in the realm of bearing prediction and fault diagnosis. Traditional non-autoregressive networks, despite their varying window sizes, face challenges in accurately interpreting instances where bearings show a transient degradation signal. This is particularly evident in situations where there is a significant increase in the amplitude of vibration signals, which later return to normal levels, as depicted in Figure \ref{fig:1a}a and \ref{fig:1a}b. At $t=0.7 \times 10^6$, the vibration signal of a bearing shows a spike indicating degradation, but it appears stable from $1.2*10^6$ to $1.8*10^6$. This leads to a challenge where relying solely on in-window vibration signals is insufficient to conclude that the bearing is in a degraded state, a situation clearly illustrated in Figure \ref{fig:1c}c, where the model's prediction briefly dips before returning to a normal value of 1.0.
   \begin{figure*}
    \centering
    \includegraphics[width=1\linewidth]{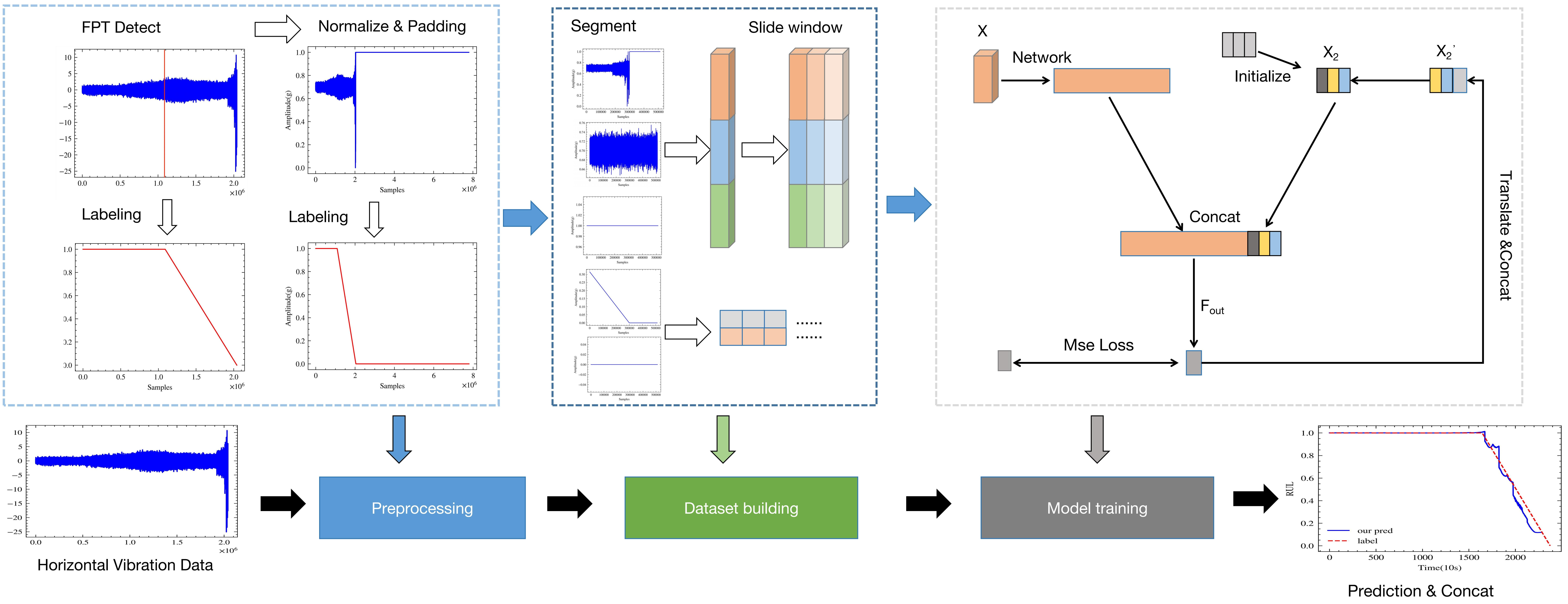}
    \caption{Proposed RUL prediction method.}
    \label{fig:flow}
\end{figure*}
To address this challenge, our paper proposes a data-driven autoregressive network structure capable of utilizing multi-input data for predictions. This model not only processes vibration signals as inputs but also incorporates previously predicted health indicator (HI) values through feature fusion, outputting current window HI values. With autoregressive iterations, the model acquires a global receptive field, allowing it to remember early degradation signals even in later stages of degradation. The workflow of the model, depicted in Figure \ref{fig:flow}, involves normalizing all data and padding it to a uniform length to ensure equal segment lengths during segmentation. This is followed by segmenting and applying a sliding window to create a dataset for model training. Once trained, the model concatenates predictions from each segment to form the final prediction outcome.
The main contributions of this paper can be summarized as follows:

\begin{enumerate}

\item{The paper introduces a novel multi-input autoregressive model tailored for predicting the remaining useful life (RUL) of bearings. This model uniquely integrates vibration signal inputs with previously predicted health indicator (HI) values, utilizing feature fusion to output HI values for the current window.}

\item{To tackle the issue of error accumulation in autoregressive networks, we adopt a segmentation approach. This method involves dividing the complete bearing data into multiple independent samples, each trained separately.}

\item{The paper proposes an innovative multiple training process method to address initial point prediction deviations. Specifically, it involves selecting a subset of samples from each segment and dividing them into parts for multiple training iterations. During these iterations, the model is updated without altering the label values until a specified number of training cycles is completed.}

\item{The proposed method was empirically validated on the PHM dataset, demonstrating remarkable generalization capabilities. Compared to other backbone networks using the same autoregressive approach, traditional autoregressive models, and non-autoregressive networks, our model significantly leads in terms of root mean square error (RMSE) and Score metrics, indicating a substantial improvement over existing methods.}
\end{enumerate}

\section{preliminary}
\subsection{Autoregressive principle}
 The Autoregressive (AR) model serves as a fundamental tool in time series data analysis, particularly for delineating the relationship between a variable and its own lagged observations. Within the scope of AR models, the first-order autoregressive model (AR(1)) and the second-order autoregressive model (AR(2)) are two commonly used orders. The AR(1) model encapsulates the relationship between the current value and its immediate predecessor, while the AR(2) model extends this relationship to include the influence of the two preceding time points. These models are mathematically represented to quantify the impact of past observations on the current value using autoregressive coefficients and incorporate an error term to account for unexplained variability. Furthermore, the AR model can be generalized to an AR(p) model, where 'p' denotes the order of the model. This generalization allows for a more flexible approach to capturing dependencies across different time points, with its mathematical formulation expressed as follows:
\begin{equation}
    X_t=c+\sum_{i=1}^p\phi_iX_{t-i}+\epsilon_t
\end{equation}
Here, $X_t$ represents the series value at time point $t$; $c$ is a constant term (which may be zero); $p$ denotes the order of the model, indicating the number of past consecutive observations included up to the current value; $\phi_i$ are the autoregressive coefficients, quantifying the impact of past observations on the current value; and $\epsilon_t$ is the error term, representing random disturbances.

Leveraging the advantageous properties of autoregressive models, we can effectively utilize the historical health indicator (HI) data to guide the prediction of HI values for the current window. This approach differs fundamentally from models like recurrent neural networks (RNNs), which perform autoregressive operations internally on sequential information within a single data instance. Instead, we have devised a training paradigm that conducts autoregressive operations within a window, allowing for predictions across the entire lifecycle of the equipment. This window-based autoregressive training mode enables a comprehensive and holistic approach to lifecycle prediction, distinct from traditional sequence-based methods.

\subsection{Dataset Construction}
In the ideal scenario, each bearing's entire vibration signal dataset could serve as a sample for autoregressive training and prediction. However, in practice, due to a severe shortage of bearing data, the model lacks robust generalization capability, making it challenging to predict the entire vibration signal. A more practical approach is to segment the signal, treating each segment as an independent sample for autoregressive training. By concatenating the prediction results of each segment, we can forecast the entire vibration signal.
For a complete vibration signal dataset D of a bearing, we divide it evenly into $n$ segments, each of length $m$:
\begin{equation}
\mathbf{D}=
\left[\begin{array}{ccccc}
v_1 & v_2 & v_3 & \ldots & v_m \\
v_{m+1} & v_{m+2} & v_{m+3} & \ldots & v_{2m} \\
\vdots & \vdots & \vdots & \ddots & \vdots \\
v_{(n-1)m+1} & v_{(n-1)m+2} & v_{(n-1)m+3} & \ldots & v_{(n-1)m}
\end{array}\right]
\end{equation}
Here  $v_n$ represents the vibration signal data at each moment, with a dimension of $N_{feature}$. 

For each segment, we use a sliding window of size $k$, meaning the data corresponding to the sampling points $0-k$ are taken as an input sample. Since the vibration data has two channels, horizontal and vertical, the input data x has a dimension of $2k*N_{feature}$. For the initial window, the corresponding label is designated as the $(k+1)^{th}$ value of the remaining useful life (RUL) curve. Subsequently, this window is shifted one time unit forward along the sampling time axis, and the process is repeated to generate a series of input samples. Within each segment, this methodology yields $m-k$ samples along with their corresponding labels. It is important to note that the last window is excluded from data construction due to the absence of a subsequent label.

To ensure uniformity in the length of segments post the application of a sliding window, padding is applied to the normalized vibration data and labels. Specifically, let $l$ denote the maximum number of samples for any bearing across all datasets. Our objective is to pad the number of samples for each bearing after the sliding window operation to the nearest multiple of $n$ that is also divisible by 100. Mathematically, the padding length, denoted as $l_{pad}$, can be expressed as follows:

\begin{equation}
\begin{aligned}
len&=l-k \\
l_f&=\left\lceil\frac{len}{\text{LCM}(100,n)}\right\rceil\times\text{LCM}(100,n)\\
l_{pad}&=l_f-len \\
\end{aligned}
\end{equation}
where LCM(100,n) represents the least common multiple of 100 and $n$.

\subsection{Network Structure}
In our experiment, a conventional 1D convolutional neural network (1DCNN) architecture was employed, which remarkably outperformed other state-of-the-art (SOTA) networks in terms of performance. The specific structure of the model is detailed in Table \ref{tab:cnn1d}. The backbone of the model consists of five convolutional blocks, each comprising a convolutional layer, a batch normalization (BN) layer, a rectified linear unit (ReLU) layer, and a max-pooling layer. All convolutional layers in the network utilize kernels of size 3x1 with a stride of 1 and padding of 1, maintaining the size of the features. The stride of the third pooling layer is set to 4 to reduce the number of parameters. Upon input, the data from all windows are concatenated along the channel dimension, resulting in a 2k-dimensional input. The input labels undergo a transformation through a 1x1 convolution for feature matching, and subsequently, the transformed labels are concatenated with the input data along the channel dimension.
\begin{figure}
    \centering
    \includegraphics[width=1\linewidth]{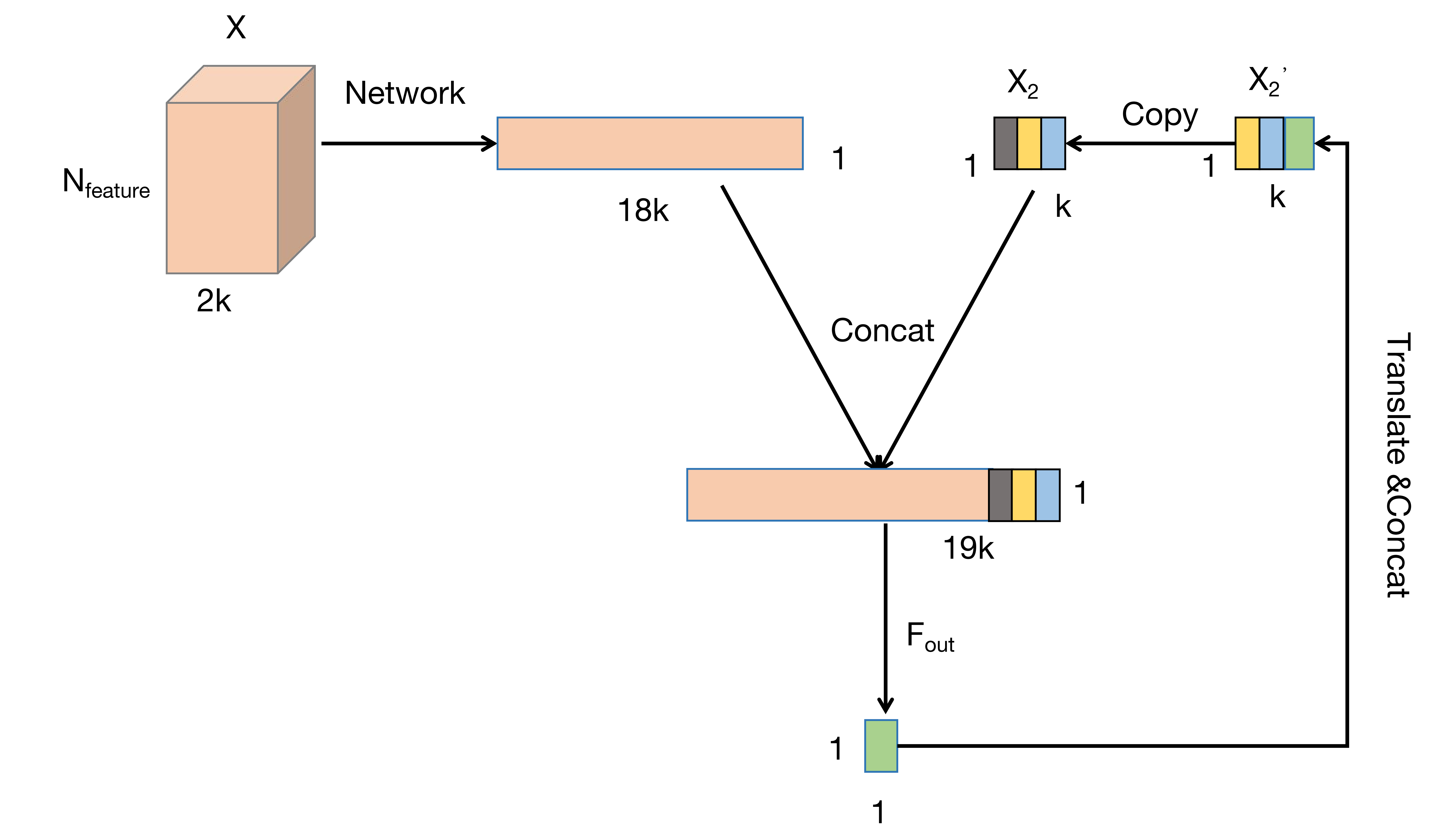}
    \caption{Autoregressive working principle diagram}
    \label{fig:method}
\end{figure}
The operational principle of the model is depicted in Figure \ref{fig:method}. On the left side, the input comprises vibration signals that first undergo feature extraction through the network. These extracted features are then concatenated with features from the right side branch and processed through the $F_{out}$ operation. This operation involves sequentially feeding the combined features into a fully connected layer, followed by a RELU layer, dropout, and another fully connected layer before producing the output. The right-side branch represents the labels corresponding to the current vibration signals. In the first epoch, this branch is initialized with the tensor of the first window of the label values. At the end of each model run, $x_2$ is shifted one unit to the left (discarding the first point of $x_2$) and then concatenated with the final prediction result. All network frameworks discussed in the paper maintain this structure, with the only variation being the replacement of the 'Network' with different backbone networks.

\begin{table}[htbp] \centering \caption{The structure of our model. \label{tab:cnn1d}}  \begin{tabularx}{\linewidth}{XXXXX} \toprule \textbf{Layer name} & \textbf{Kernels size} & \textbf{Stride} & \textbf{Kernels number} & \textbf{Output size} \\ \midrule Input & – & – & – & 2560 x 2k \\ ConvBlock1 & 1 x 3 & 1 x 1 & 32 & 2560 x 32 \\ Pooling1 & 1 x 2 & 1 x 2 & – & 1280 x 32 \\ ConvBlock2 & 1 x 3 & 1 x 1 & 32 & 1280 x 32 \\ Pooling2 & 1 x 2 & 1 x 2 & – & 640 x 32 \\ ConvBlock3 & 1 x 3 & 1 x 1 & 64 & 640 x 64 \\ Pooling3 & 1 x 4 & 1 x 4 & – & 160 x 64 \\ ConvBlock4 & 1 x 3 & 1 x 1 & 128 & 160 x 128 \\ Pooling4 & 1 x 2 & 1 x 2 & – & 80 x 128 \\ Convolution5 & 1 x 3 & 1 x 1 & 18k & 80 x 18k  \\ Pooling5 & 1 x 2 & 1 x 2 & – & 40 x 450  \\ \bottomrule \end{tabularx}  \end{table}

\subsection{Multiple Training}
\begin{figure}
    \centering
    \includegraphics[width=0.45\textwidth]{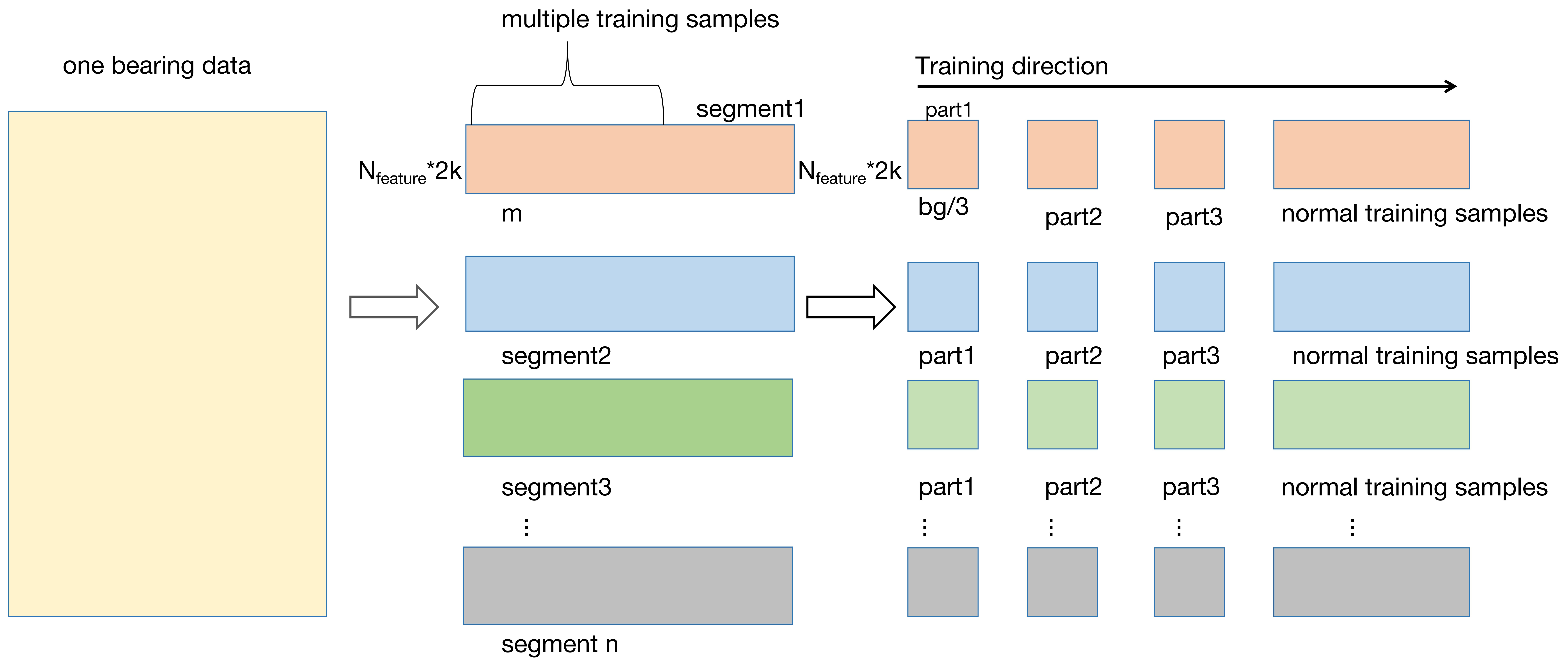}
    \caption{Data block diagram}
    \label{fig:particle}
\end{figure}

In autoregressive prediction, an initial forecasting error at early time points can lead to a cascading effect of errors in subsequent predictions, a phenomenon known as error accumulation. To address this challenge, our paper proposes a strategic approach involving multiple training iterations at the initial time points to minimize error, followed by a single standard training iteration for later time points. Specifically, for each segment, the first 'bg' samples are divided into $z$ equal-length parts, where typically $z$ is set to 3, as shown in Figure \ref{fig:particle}. Each part undergoes a number of training iterations, denoted as $z_j$, which can vary for different parts. After each iteration, the model parameters are updated, but the value of  $x_2$ is not immediately revised until $z_j$ iterations are completed. For the samples in the latter part of each segment, from $v_{bg}$ to $v_m$, a standard single training iteration is applied. It is also feasible to set different $zj$ values for different epochs, as in later epochs, the model is often capable of accurately predicting within the current window, thereby reducing the need for multiple iterations. The specifics of this algorithm are detailed in Algorithm  \ref{alg:multiple}.

\begin{algorithm}
\caption{Multiple training iterations}
\label{alg:multiple}
\begin{algorithmic}[1]
\Require x, x2, y, model
\State m: length of each segment
\For{step=1, ..., m}
    \State train\_iters=find\_train\_iters(step)
    \For{j = 1,...,train\_iters}
    \State y\_prediction = model(x, x2)
    \State loss = criterion(y\_prediction, y)
    \State loss.backward()
    \EndFor
    \State x2 = concat(x2[:, 1:], y\_prediction, dim=1)
  \EndFor
\end{algorithmic}
\end{algorithm}

\section{Experiments Overview}
\subsection{Dataset Introduction}
\begin{figure}
    \centering
    \includegraphics[width=1\linewidth]{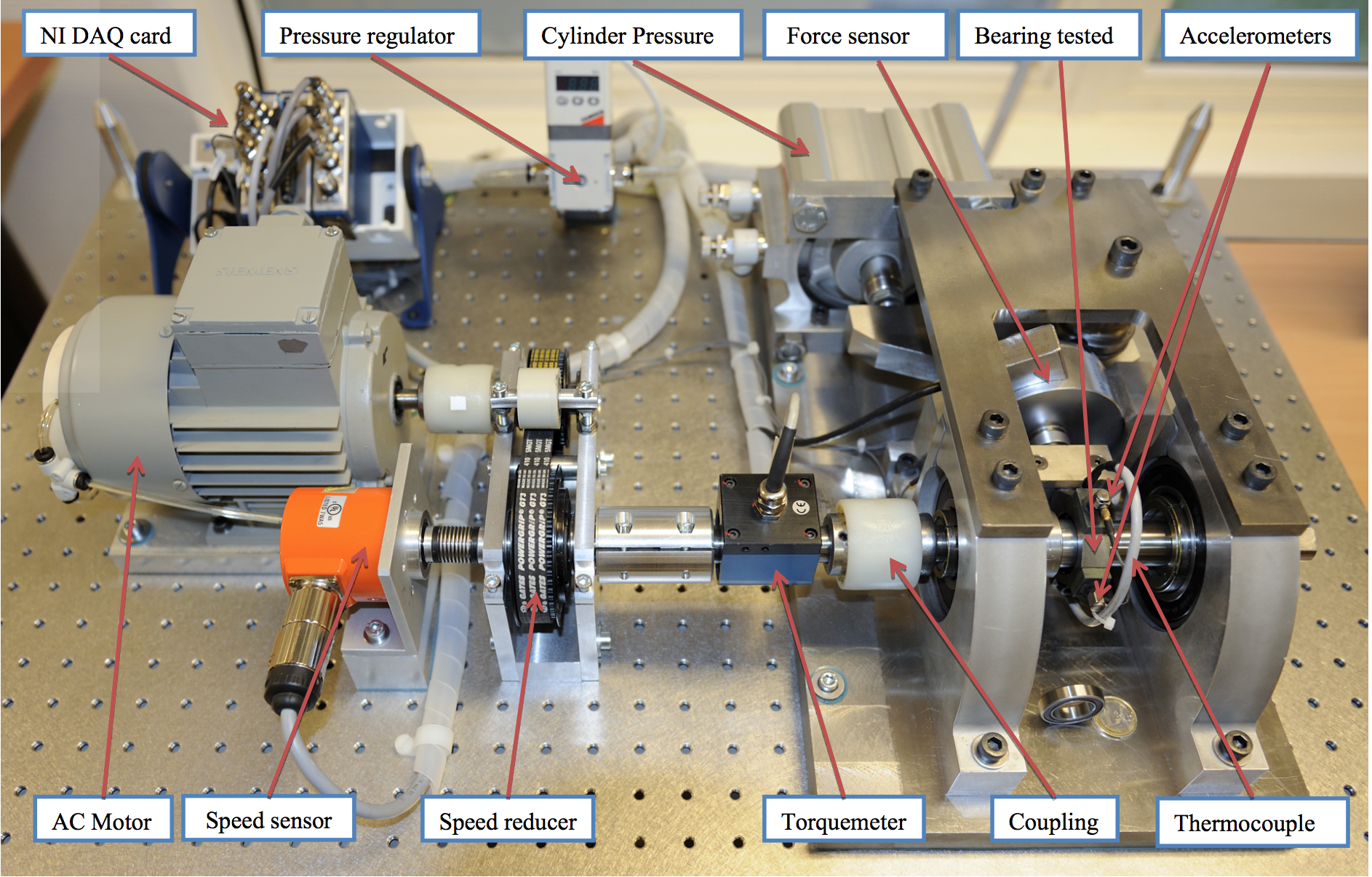}
    \caption{Overview of PRO-NOSTIA}
    \label{fig:dataset}
\end{figure}

In this paper, we utilized a dataset that records the entire process of bearing performance from normal operation to failure for method validation \cite{dataset}. This dataset was originally introduced in the IEEE PHM2012 Data Challenge. The data in this dataset were captured using the PRO-NOSTIA test rig, as illustrated in the Figure \ref{fig:dataset}. The data collection involved two accelerometers installed in horizontal and vertical orientations, respectively. These accelerometers were set to sample at a frequency of 25.6 kHz, performing data collection every 10 seconds, with each collection lasting 0.1 seconds, resulting in 2560 data points per sample. To ensure the safety of the tests, if the amplitude of vibration data exceeded 20 g, the test would be immediately terminated. The time of bearing failure was determined based on the moment this amplitude threshold was exceeded. Our predictive framework evaluated data from continuous operation to failure under two different conditions, as shown in Table \ref{tab:dataset}.

\begin{table*} 
\caption{ Bearing full life under two different working conditions.} 
\centering 
\begin{tabularx}{\textwidth}{XXXXX} 
\toprule 
\textbf{Operating Conditions} & \textbf{Working Load} & \textbf{Rotating Speed (r/min)} & \textbf{Bearing Number} & \textbf{Actual Life (s)} \\ 
\midrule \multirow{7}{*}{condition one} & \multirow{7}{*}{4000 N} & \multirow{7}{*}{1800} & Bearing1-1 & 28,030 \\ & & & Bearing1-2 & 8,710 \\ & & & Bearing1-3 & 23,750 \\ & & & Bearing1-4 & 14,280 \\ & & & Bearing1-5 & 24,630 \\ & & & Bearing1-6 & 24,480 \\ & & & Bearing1-7 & 22,590 \\ \midrule \multirow{5}{*}{condition two} & \multirow{5}{*}{4200 N} & \multirow{5}{*}{1650} & Bearing2-1 & 9,110 \\ & & & Bearing2-2 & 7,970 \\ & & & Bearing2-3 & 19,550 \\ & & & Bearing2-4 & 7,510 \\ & & & Bearing2-5 & 23,110 \\ \bottomrule \end{tabularx}  
\label{tab:dataset} 
\end{table*}

\begin{figure*}[htbp]
    \centering
    \begin{tabular}{cc}
        \includegraphics[width=0.45\textwidth]{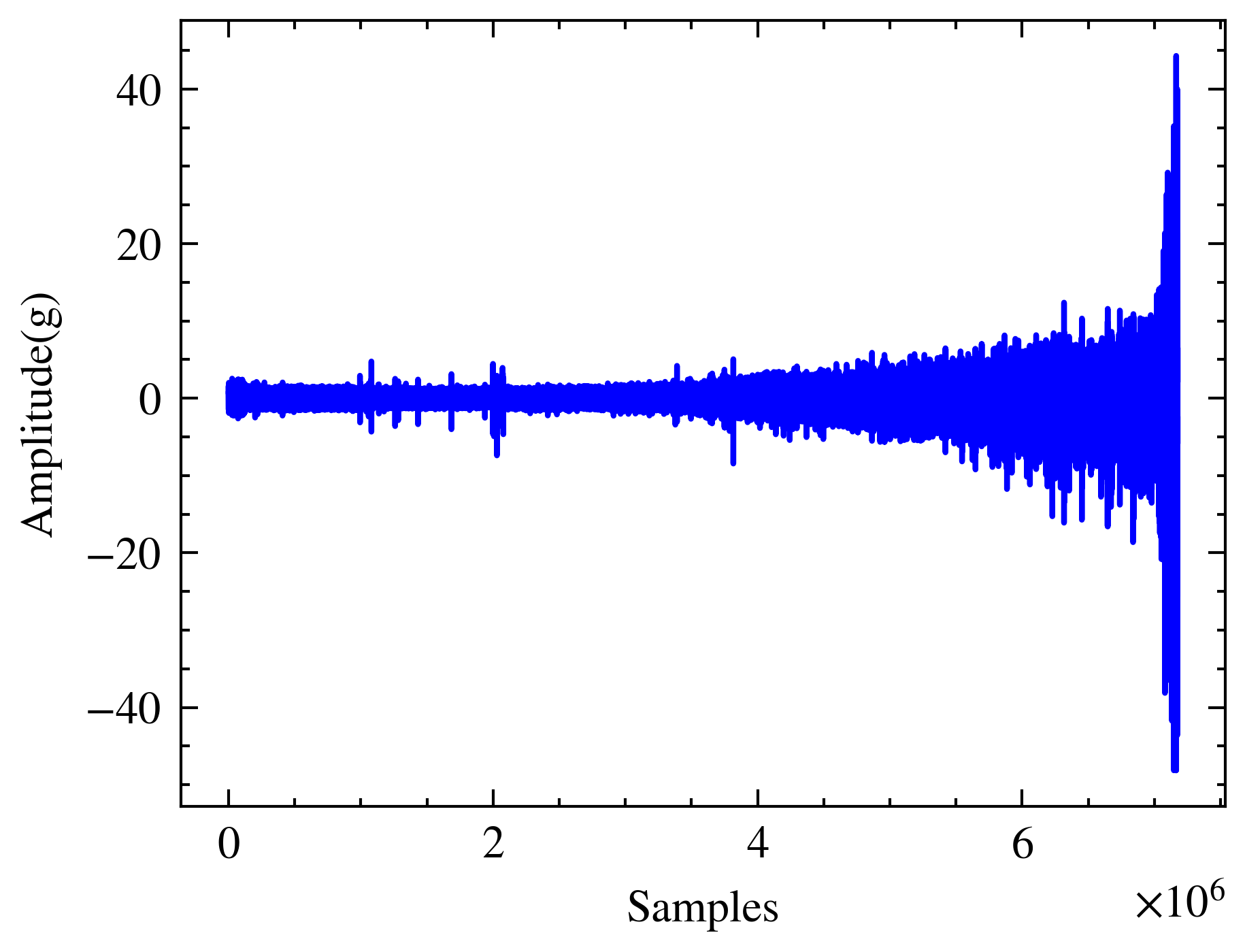} & \includegraphics[width=0.45\textwidth]{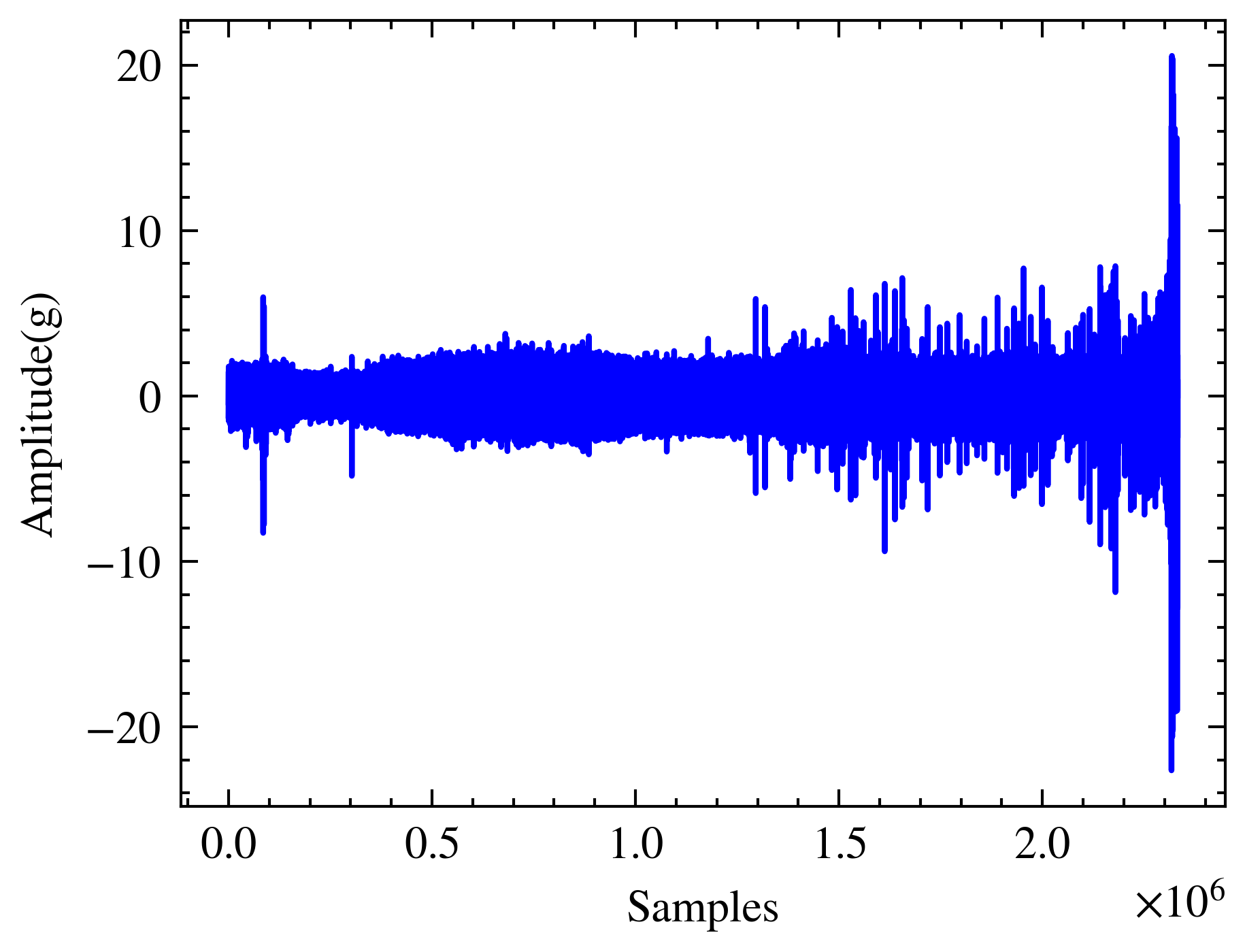} \\
        (a) bearing 1-1 & (b) bearing 1-3 \\
    \end{tabular}
    \caption{The horizontal vibration signals of PHM2012 dataset. (a) bearing 1-1. (b) bearing 1-3.}
    \label{fig:bearing}
\end{figure*}
\subsection{Parameter Setting}
During the training process of the predictive model, the AdamW optimizer was used to minimize the loss function value, with a learning rate set at 0.0008 and a batch size of 45. The network was trained over 6 epochs. If a two-dimensional convolutional network is used, the data will first undergo continuous wavelet transform (CWT) processing to be converted into two-dimensional images.The data for each bearing was augmented to 3000 samples, with vibration signals filled using 1, and labels filled using 0. All data were divided into 15 segments, each of length 200. In each training session, the complete data from all three bearings were input, which is why the batch size was set to 45. The results presented in this paper are the average of five independent runs using random seeds of 15, 25, 35, 45, and 55, ensuring consistency and repeatability. All experiments were conducted on an NVIDIA GeForce 3090 (24GB) GPU and implemented using the PyTorch framework.

Currently, in the domain of remaining useful life (RUL) prediction, two primary methods for label incorporation are prevalent, as detailed in references \cite{ref33,ref34,ref35,ref36,ref37}. The first approach employs a linear function where the RUL value linearly decreases from 1 to 0. However, this method does not account for the operational reality of bearings, wherein the RUL should remain constant during normal operational phases. The alternative approach utilizes a piecewise function to more accurately mimic the degradation trend of bearings. In this model, the health indicator (HI) of a bearing remains constant at 1 from 0 to the first prediction time (FPT) and subsequently undergoes a linear decline. For determining the FPT in the piecewise label, we adopt the methodology proposed by Li et al.\cite{fpt}, which is based on the $3\sigma$ technique, as shown in Table \ref{tab:segment}.

\begin{table}[htbp]
    \caption{ FPT of PHM2012 dataset under condition 1 and condition 2. \label{tab:segment} }
    \begin{tabularx}{\linewidth}{XXX}
        \toprule
        \textbf{Bearings}  & \textbf{FPT}   &\textbf{Actual Life(s)}  \\
        \midrule
        B1-1   & 11420 & 28030 \\
        B1-2   & 8220 & 8710 \\
        B1-3   & 9600 & 23750 \\
        B1-4  & 10180 & 14280 \\
        B1-5  & 24070 & 24630 \\
        B1-6  & 16270 & 24480 \\
        B1-7  & 22040 & 22590 \\
        B2-1  & 2100  & 9110 \\
        B2-2  & 2460  & 7970 \\
        B2-3  & 2730  & 19550 \\
        B2-4  & 3720  & 7510 \\
        
        \bottomrule
        \end{tabularx}
\end{table}
The methodology employed for data division into training and test sets involves selecting one bearing from the seven available for testing, while utilizing the remaining six for training purposes. Specifically, in this paper, bearings 1-1 and 1-3 are primarily used as test sets. Their lifecycle data are visualized in Figure \ref{fig:bearing}. Consequently, for testing with bearing 1-1, bearings 1-2 to 1-7 are used for training. Similarly, when testing with bearing 1-3, bearings 1-1, 1-2, and 1-4 to 1-7 serve as the training set.

\subsection{Metrics}

To comprehensively evaluate the accuracy of our model's predictions, two widely recognized metrics are employed: root mean square error (RMSE) and mean absolute error (MAE). These metrics are chosen for their ability to provide a clear and quantitative measure of prediction errors. The RMSE is particularly useful in emphasizing larger errors, as it squares the residuals before averaging, thereby giving more weight to larger error margins. On the other hand, MAE offers a direct average of absolute errors, making it intuitive and easy to interpret. The formulas for RMSE and MAE are as follows: 
\begin{equation}\text{RMSE}=\sqrt{\frac{1}{N}\sum_{i=1}^{N}\left(\hat x_i-x_{i}\right)^{2}}\end{equation}
\begin{equation}\text{MAE}=\frac{1}{N}\sum_{i=1}^{N}|\left(\hat x_i-x_{i}\right)|\end{equation}
where $\hat x_i$ denotes actual RUL of the $i^{th}$ predictive object, $x_i$ signifies the predicted RUL of the $i^{th}$ predictive object and $N$ denotes the number of the samples. 

In addition to the standard metrics of RMSE and MAE, our evaluation framework also incorporates the scoring metric recommended by the PHM2012 official challenge,  detailed in Equation (6). This scoring metric is particularly tailored to aggregate the deviation between predicted and actual values, with a specific emphasis on penalizing positive errors. The rationale behind this emphasis is grounded in practical considerations; in real-world production scenarios, delayed predictions (positive errors) often pose a more significant risk compared to early predictions (negative errors).
\begin{equation}
\begin{aligned}
E_{i}&=\hat x_{i}-x_{i}\\
A_i&=\begin{cases}
\exp^{-(\frac{E_{i}}{13})} -1
&\text{ if }E_{i}\leq0\\
\exp^{(\frac{E_{i}}{10})}  -1
&\text{ if }E_{i}>0 
\end{cases} \\
Score&=\sum_{i=1}^{N}(A_{i}) 
\end{aligned}
\end{equation}
where $\hat x_i$ denotes actual RUL of the $i^{th}$ predictive object, $x_i$ signifies the predicted RUL of the $i^{th}$ predictive object, and $N$ denotes the total number of samples. 

\section{Experiments Details}
\subsection{Ablation Study}
\begin{figure*}[htbp]
   \begin{tabular}{ccc}
        \includegraphics[width=0.33\textwidth]{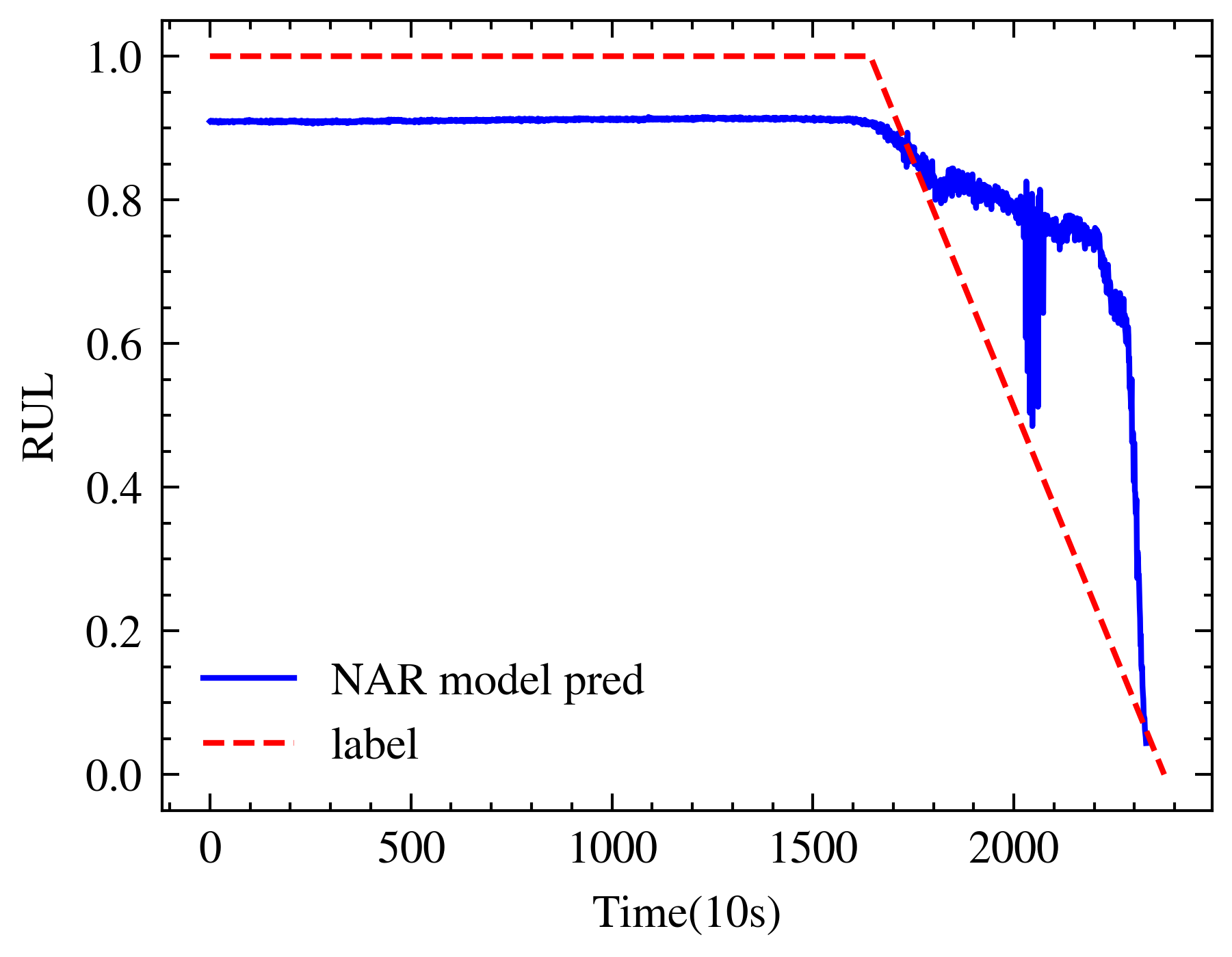} & \includegraphics[width=0.33\textwidth]{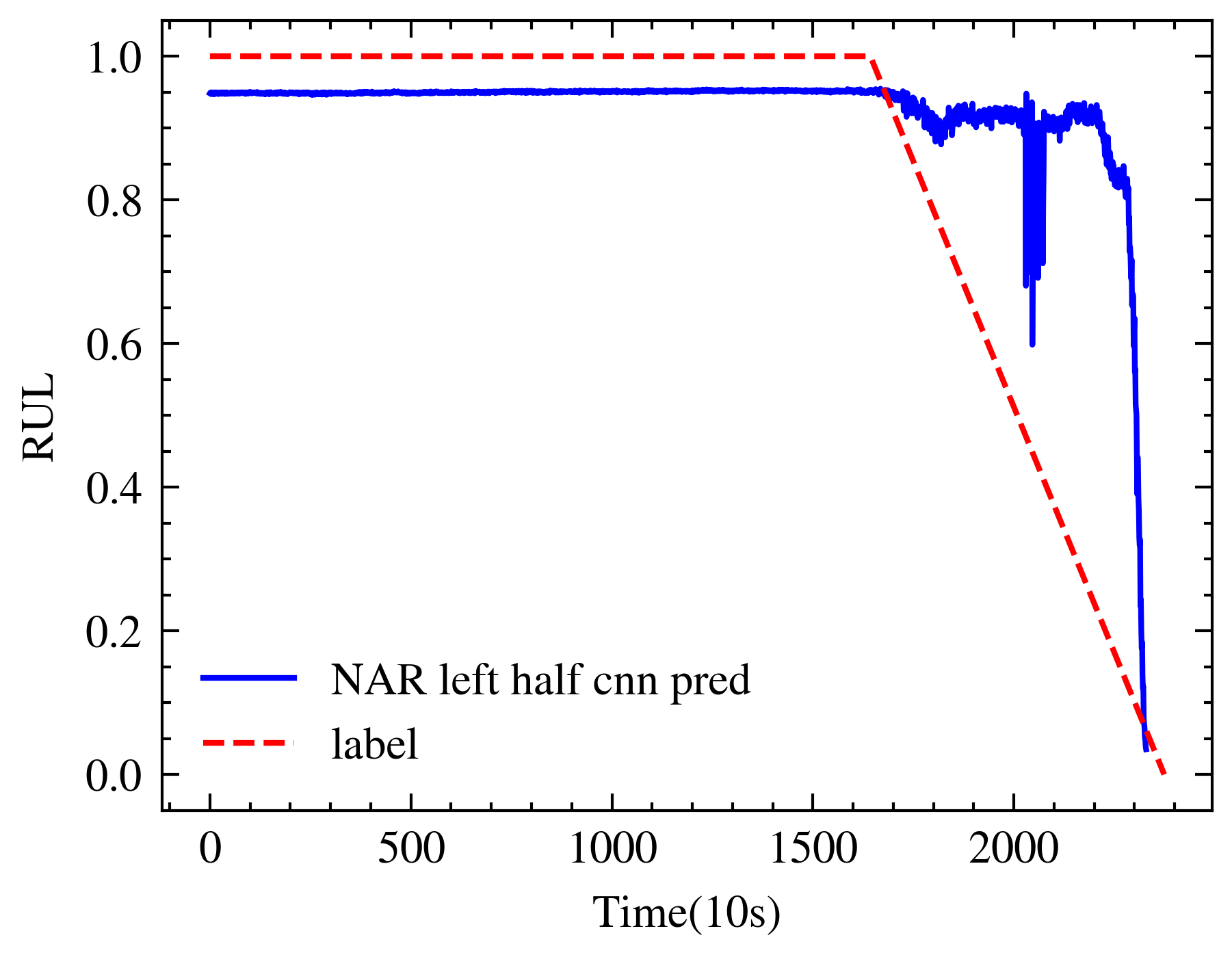} &
        \includegraphics[width=0.33\textwidth]{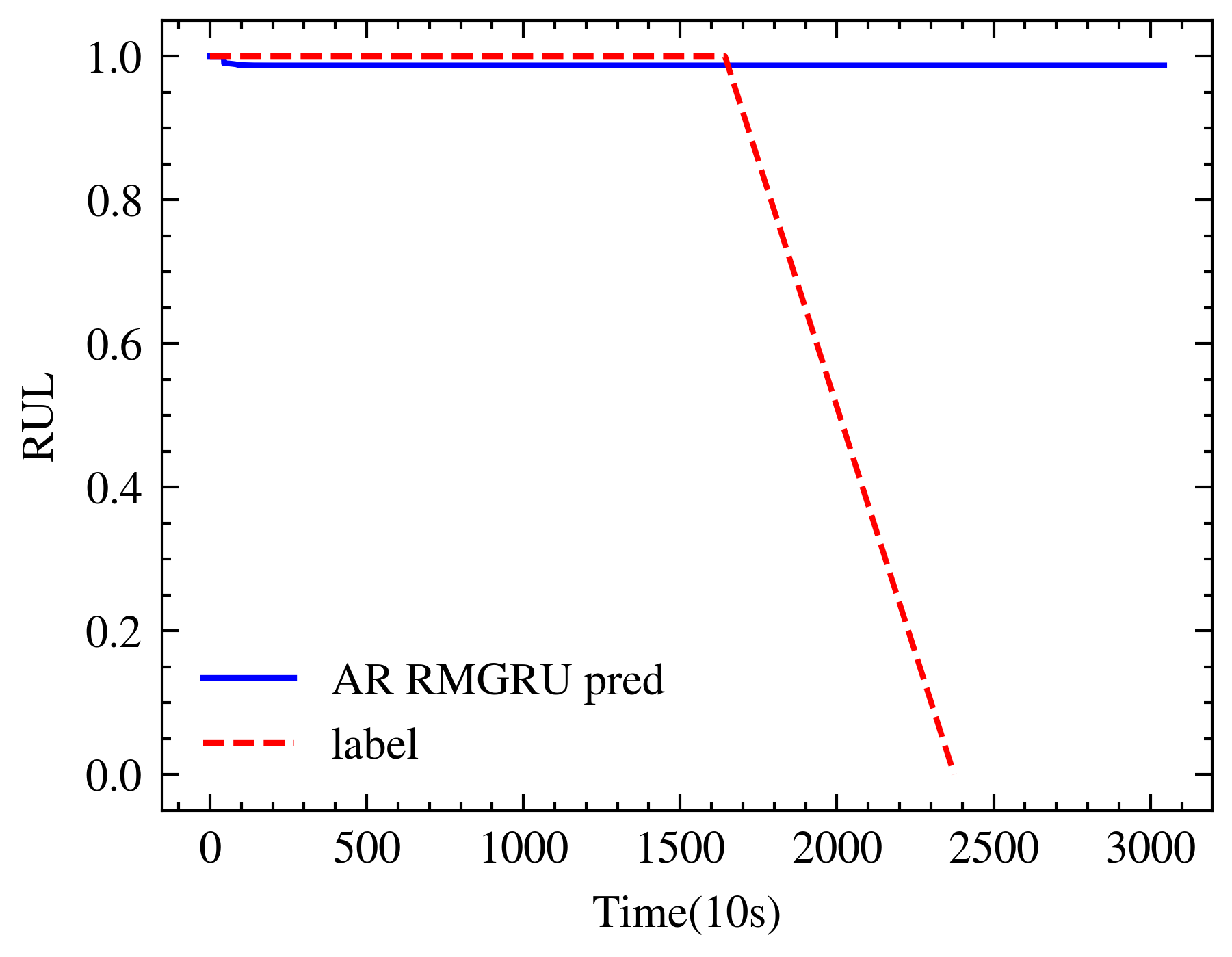} 
        \\
        (a)  & (b) & (c) \\
    \end{tabular}
    \caption{Ablation experiment on bearing 1-3. (a) Our Model Employing Non-Autoregressive Training. (b) The Left Half of Our Model Showcasing the CNN Network Utilizing Non-Autoregressive Training. (c)  RMGRU Network Utilizing Autoregressive Training.}
    \label{fig:ablate-com}
\end{figure*}

\begin{figure*}[htbp]
   \begin{tabular}{ccc}
        \includegraphics[width=0.5\linewidth]{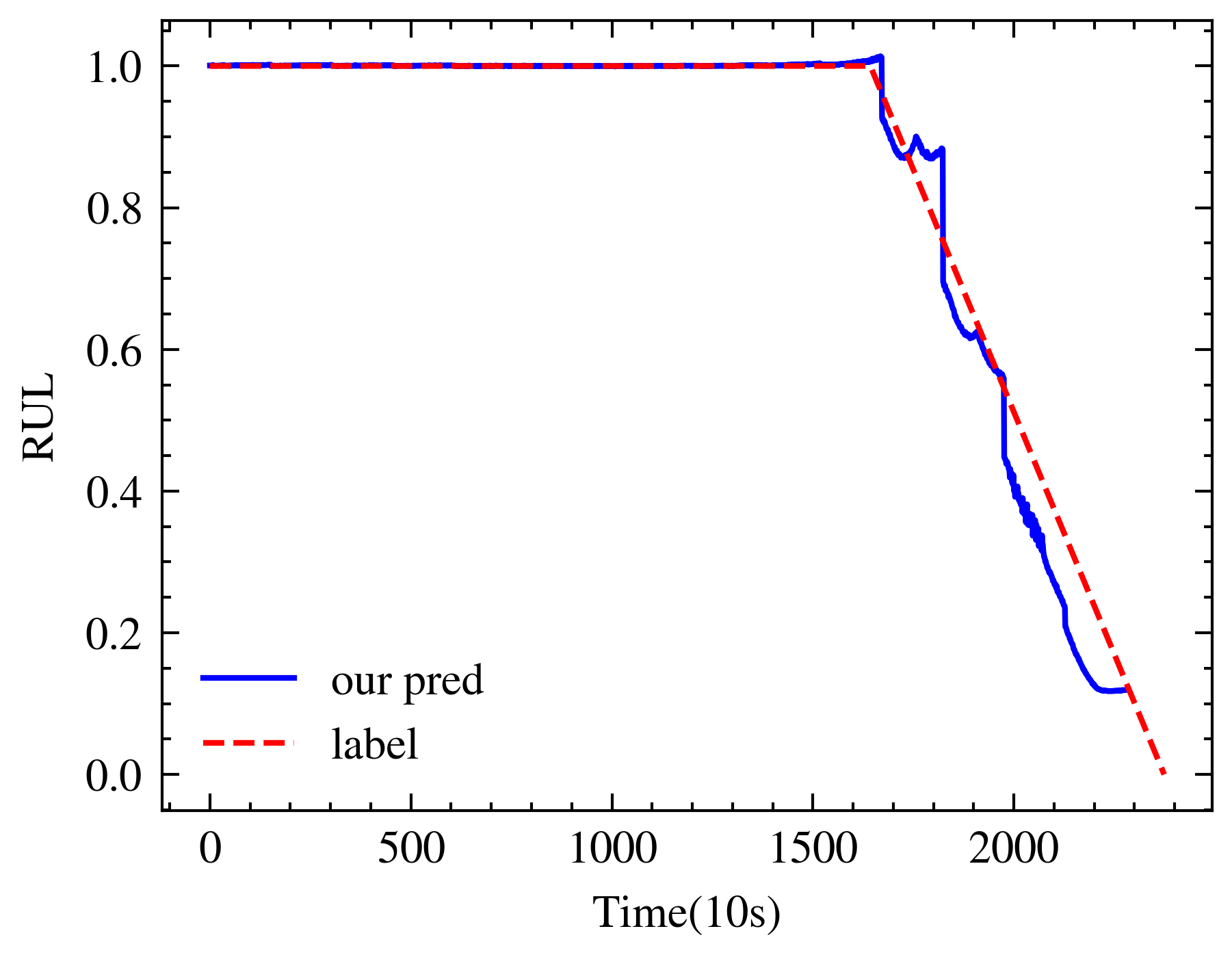} & \includegraphics[width=0.5\linewidth]{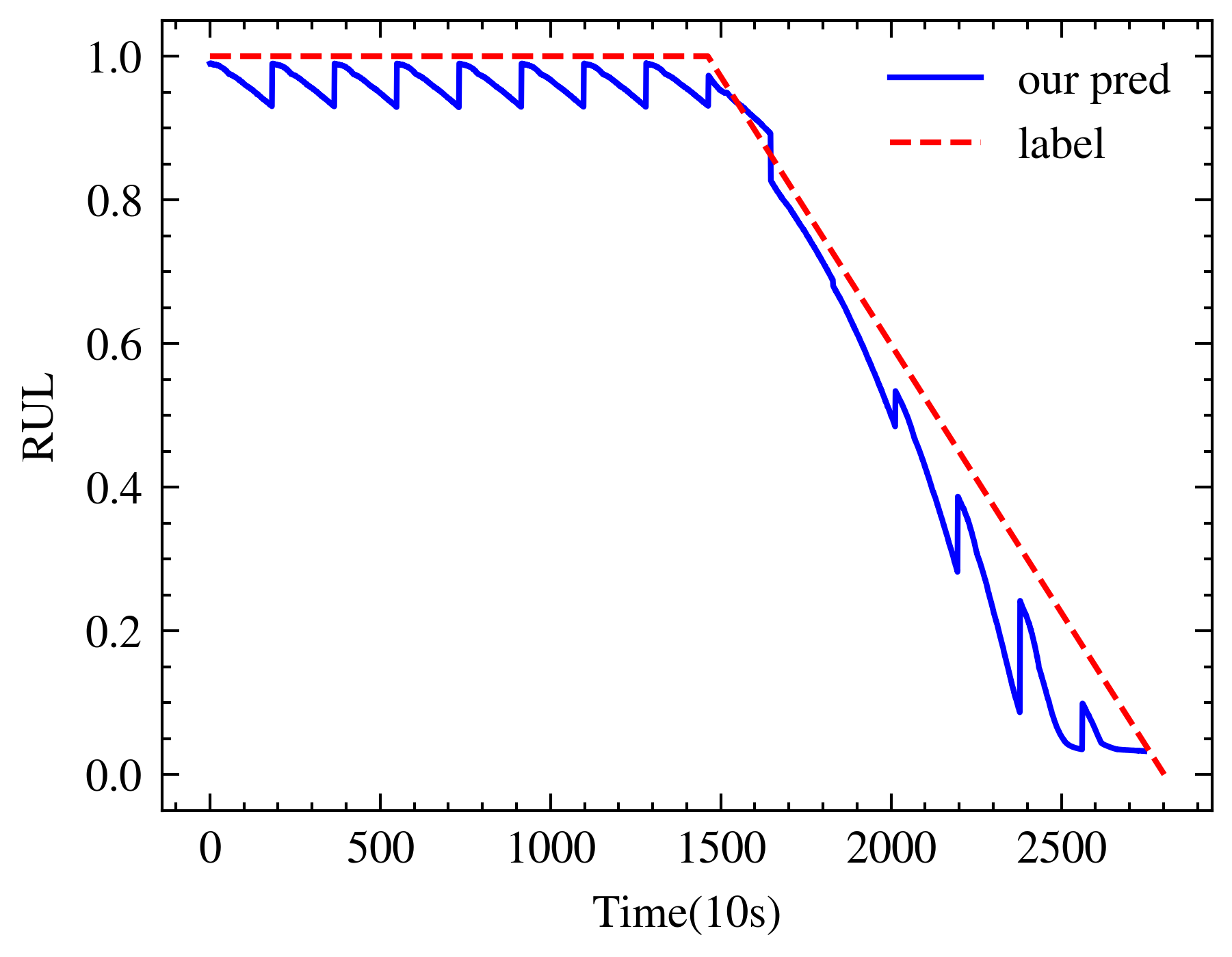} &
        \\
        (a)  & (b)  \\
    \end{tabular}
    \caption{The predictions of our model. (a) bearing1-3 (b) bearing1-1 }
    \label{fig:self-com}
\end{figure*}

To verify the effectiveness of the autoregressive algorithm, we compared it with the non-autoregressive version of the model, the left half of the model which operates in a non-autoregressive manner, and a conventional autoregressive network using only label values \cite{rmgru}. The test results for bearings 1-3 are shown in Figures \ref{fig:ablate-com} - \ref{fig:self-com}a. We observed that, apart from the autoregressive model, the performance of other models is very poor, which is attributed to the extremely limited number of training epochs, preventing other models from fitting adequately. As for the RMGRU model that only uses label values, due to its reliance solely on labels and lack of other information, the network completely lacks generalization capability, thus presenting a straight line. In contrast, our autoregressive model predicts the values of the normal phase (HI=1) very accurately. During the degradation phase, it shows some fluctuations compared to the label.
Looking at the test results for bearing 1-1, as shown in Figure \ref{fig:self-com}b, its performance is somewhat inferior to that of bearings 1-3. In the normal phase, its predicted values consistently drop from 1 to 0.95, displaying a sawtooth pattern. During the degradation phase, significant deviations appear after t=2000.
Overall, the autoregressive model, which integrates multiple inputs, shows excellent performance under low training volumes compared to single-architecture or non-autoregressive versions of the model.


\subsection{Hyperparameter Study}
\begin{figure*}[htbp]
   \begin{tabular}{ccc}
        \includegraphics[width=0.33\textwidth]{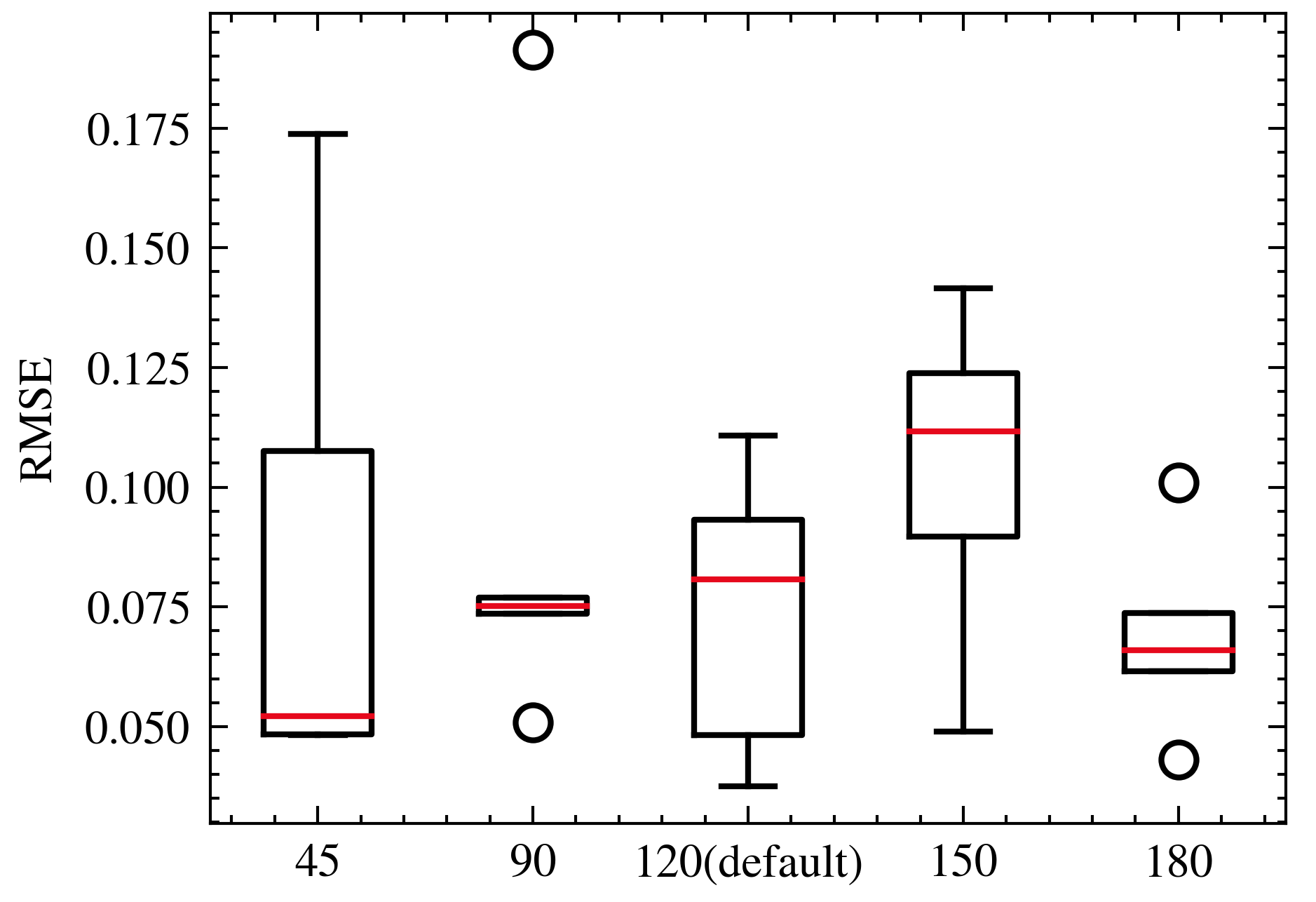} & \includegraphics[width=0.33\textwidth]{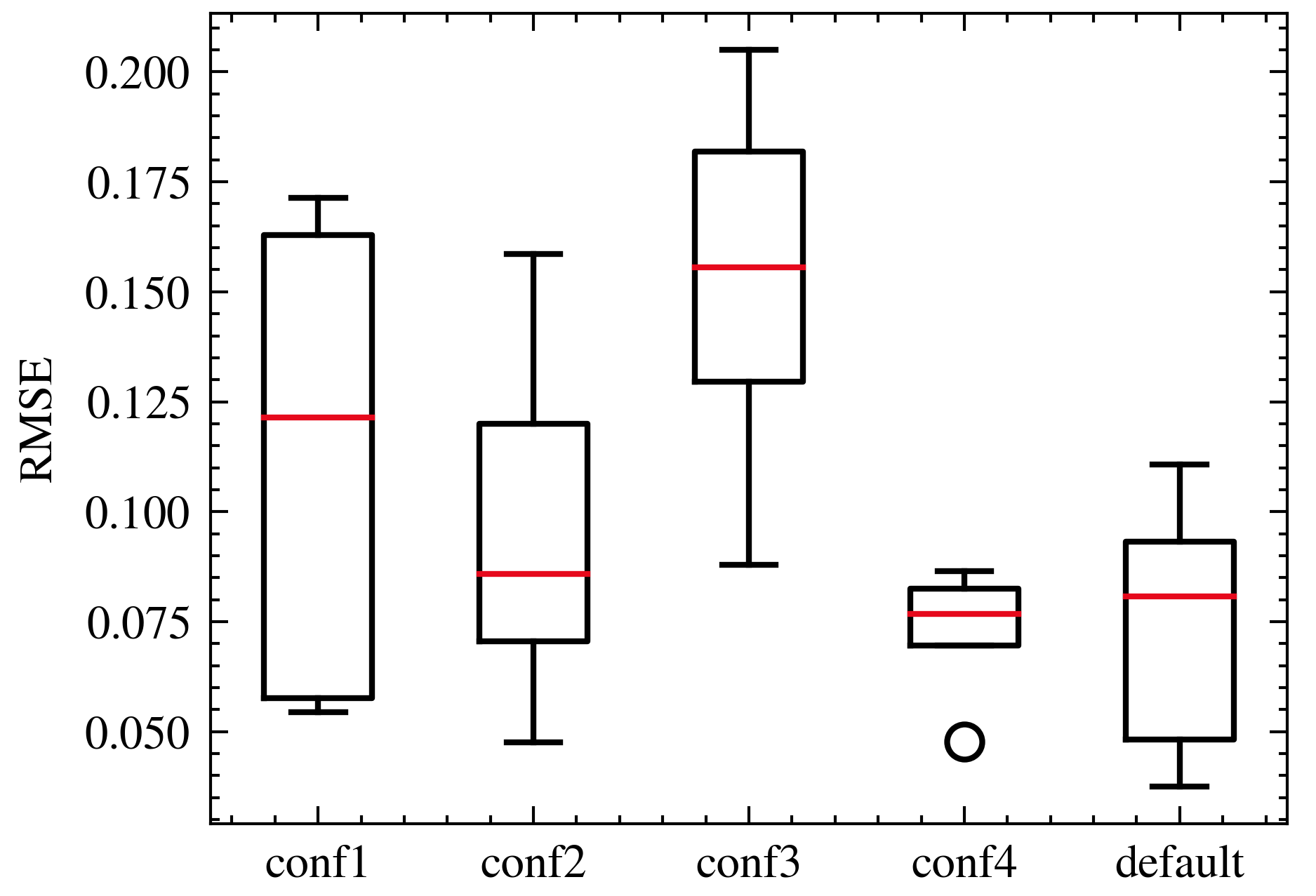} &
        \includegraphics[width=0.33\textwidth]{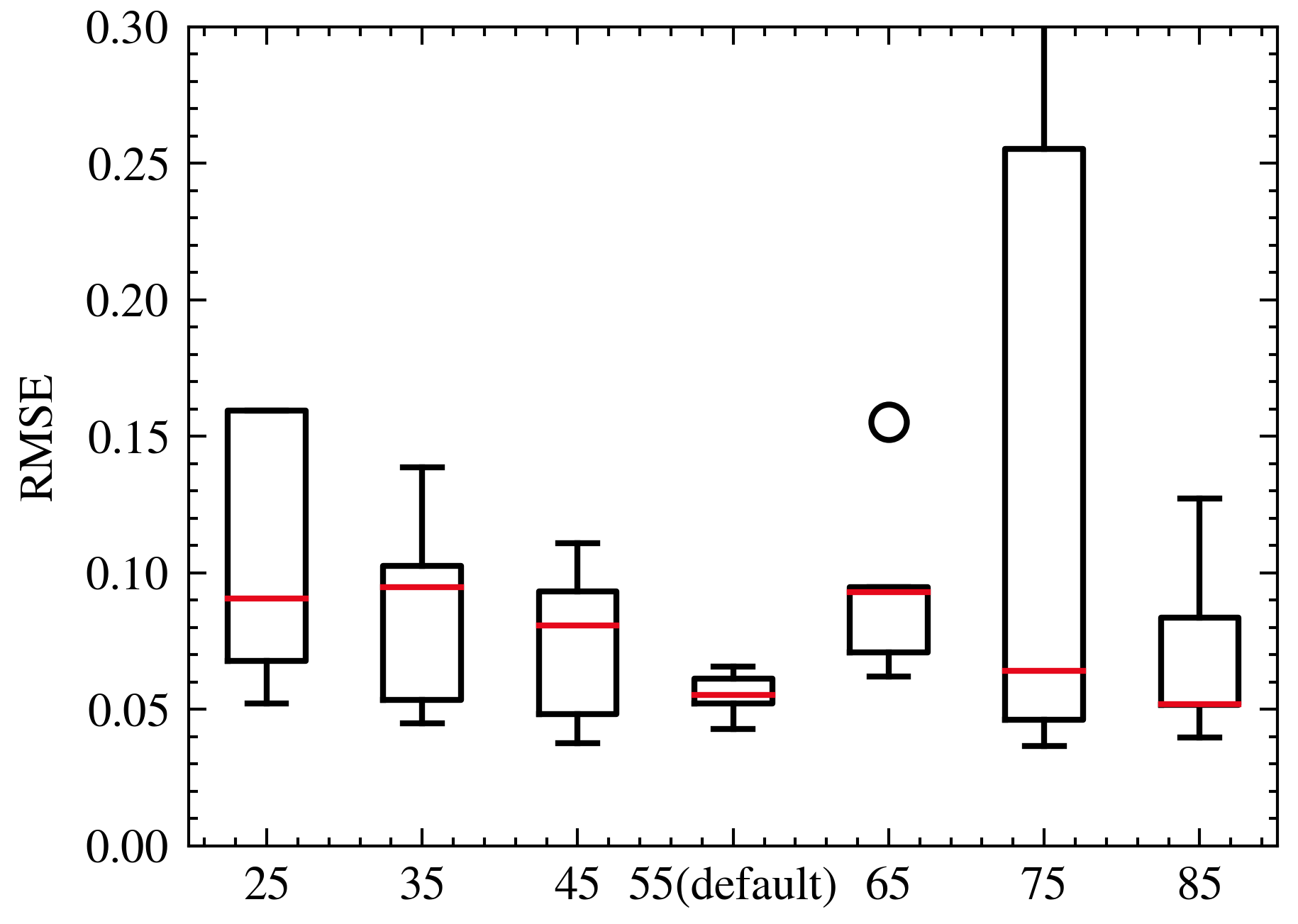} 
        \\
        (a)  & (b) & (c) \\
    \end{tabular}
    \caption{Hyperparameter evaluation results. (a) The different length of the multiple training process samples. (b) The different configurations of multiple training process. (c)  The varying lengths of window sizes.}
    \label{fig:hyper}
\end{figure*}

In this section, we conduct a series of detailed comparative experiments focusing on the three most critical hyperparameters within our method.
For different lengths of the multiple training process points, denoted as $bg$, the general trend is that as the length of the multiple training sessions increases, the RMSE first decreases and then increases, as depicted in Figure \ref{fig:hyper}a. However, $bg$=45 is an exception; it has considerable fluctuations but can yield good results with excellent initial values. At $bg$=90, except for two outliers, the overall performance is relatively stable, but not exceptionally well-performing. When $bg$=120, it achieves the best performance in most cases. Continuing to increase the length of the multiple training sessions does not yield better results and instead adds to the training cost.

\begin{table}[htbp]
\centering
\caption{The configurations of multiple training process. }
\label{tab:config}
\begin{tabularx}{\linewidth}{XX}
\toprule
\textbf{Name}   &  \textbf{Configurations}   \\
\midrule
conf1 & [3,2,2, 2,2,1, 2,1,1]    \\
conf2  & [1,1,1, 1,1,1]  \\
conf3   & [2,2,2, 2,2,2, 2,1,1]   \\
conf4 & [2,2,1, 1,1,1]   \\
default & [2,2,2, 2,2,1, 2,1,1]   \\
\bottomrule
\end{tabularx}
\end{table}
To investigate the impact of different training iterations of the multiple training process on model performance, we divided the multiple training process points into three equal parts. And we define several configurations, as the Table \ref{tab:config} shows. If the array has only six elements, the first three elements represent the multiple training iterations per part in the first epoch, and the last three elements for the other epochs. If there are nine elements, the middle three represent the second epoch's multiple training iterations. The testing results for bearing 1-3 are shown in Figure \ref{fig:hyper}b. The default configuration, differing from $conf3$ only by one less training iteration in the third part of the second epoch, showed a significant improvement in overall performance. $Conf2$, which represents the effect of not using the multiple training technique, performed better than most configurations using this technique ($conf1$, $conf3$), as seen in the figure. $Conf4$ provided stable training with minimal fluctuation but did not achieve a lower minimum value than the default. Our current exploration is not sufficient to discern a clear pattern; there is still considerable room for optimization and adjustment in the configuration of $z_j$.

For different window sizes, the overall trend observed is that with an increase in window size, the RMSE initially decreases and then increases, as illustrated in Figure \ref{fig:hyper}c. The RMSE reaches its minimum value when the window size is 55, where the training is extremely stable with minimal fluctuation. At a window size of 75, the training becomes highly unstable; despite the minimum value of RMSE remaining unchanged, many initial values can cause the network training to collapse. A window size of 85 is an exception; it has the lowest average RMSE but with significant fluctuations, and the training cost increases substantially.

\subsection{Comparison with other prediction model}
To demonstrate the superiority of our model, we selected models such as CNN2D, CBAM\cite{cbam}, SENET\cite{senet}, SKNET\cite{refsknet}, LSTM, GRU, TCN, ASTCN\cite{astcn}, and DCNN\cite{zo4} for comparison. We constructed autoregressive networks using the methods described in this paper for these networks, and all models were trained using autoregressive methods. For CNN2D, CBAM, SENET, and SKNET, we used two-dimensional images processed by CWT as input.
\begin{enumerate}
\item CNN2D: A standard two-dimensional convolutional neural network, including 4 convolutional blocks. Each block comprises a 2D convolution layer, batch 
normalization, RELU activation, and maxpooling layers.
\item LSTM: Initially passes through two convolutional blocks for dimension reduction, each block consisting of a 1D convolution layer, layer normalization, and maxpooling, forming a tensor of dimensions (270, 80). This is followed by a dropout layer, and then three LSTM layers, with a hidden size of 64 and dropout rate of 0.2.
\item GRU: Similar to LSTM, but the final layers are three GRU layers, with a hidden size of 64 and dropout rate of 0.2.
\item TCN: Similar to LSTM in initial structure, but the final layers consist of three TCN (Temporal Convolutional Network) layers, with channel sizes of 64, 360, 270, kernel size of 2, and dropout rate of 0.2.
\end{enumerate}

\begin{figure*}[htbp]
   \begin{tabular}{ccc}
        \includegraphics[width=0.5\linewidth]{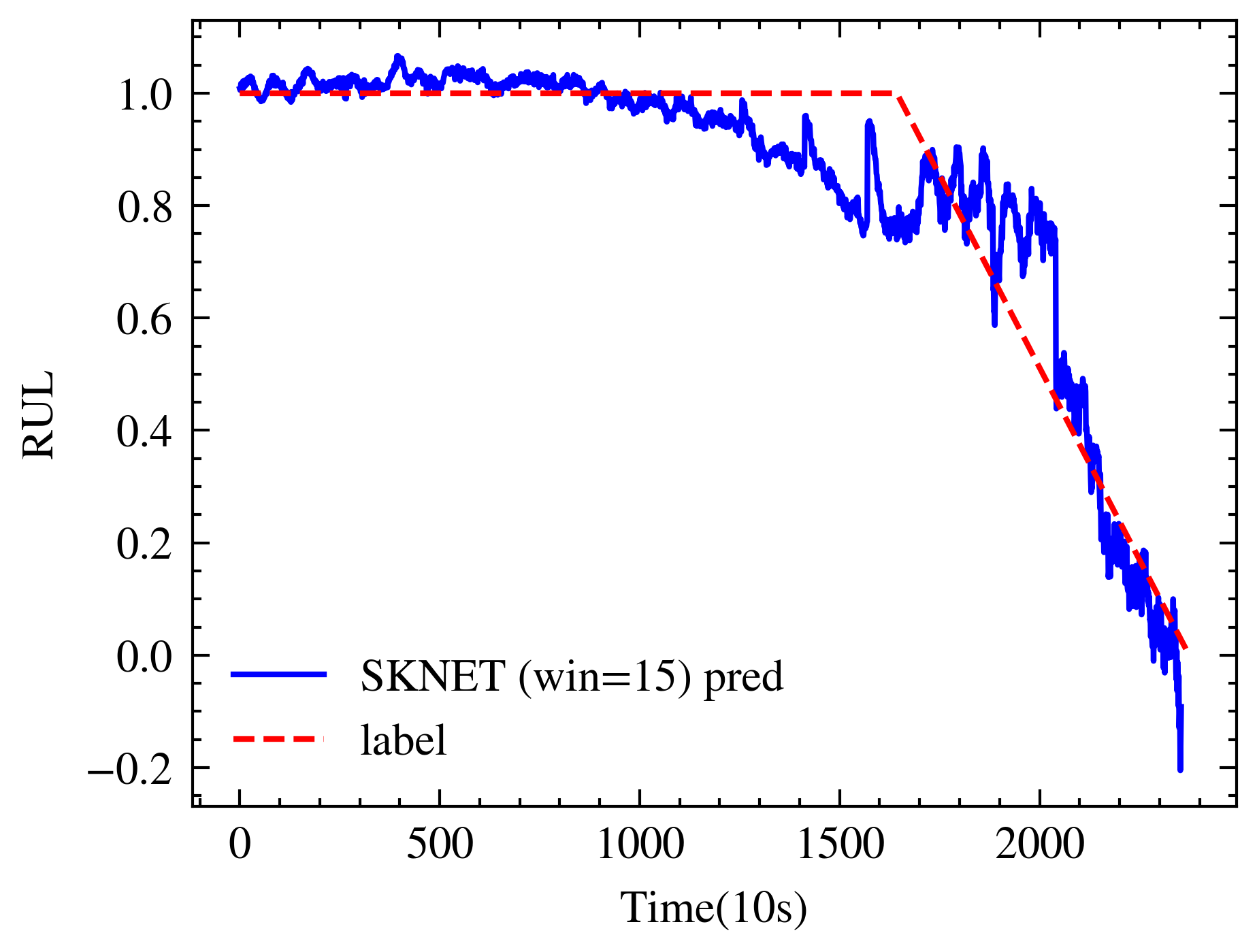} & \includegraphics[width=0.5\linewidth]{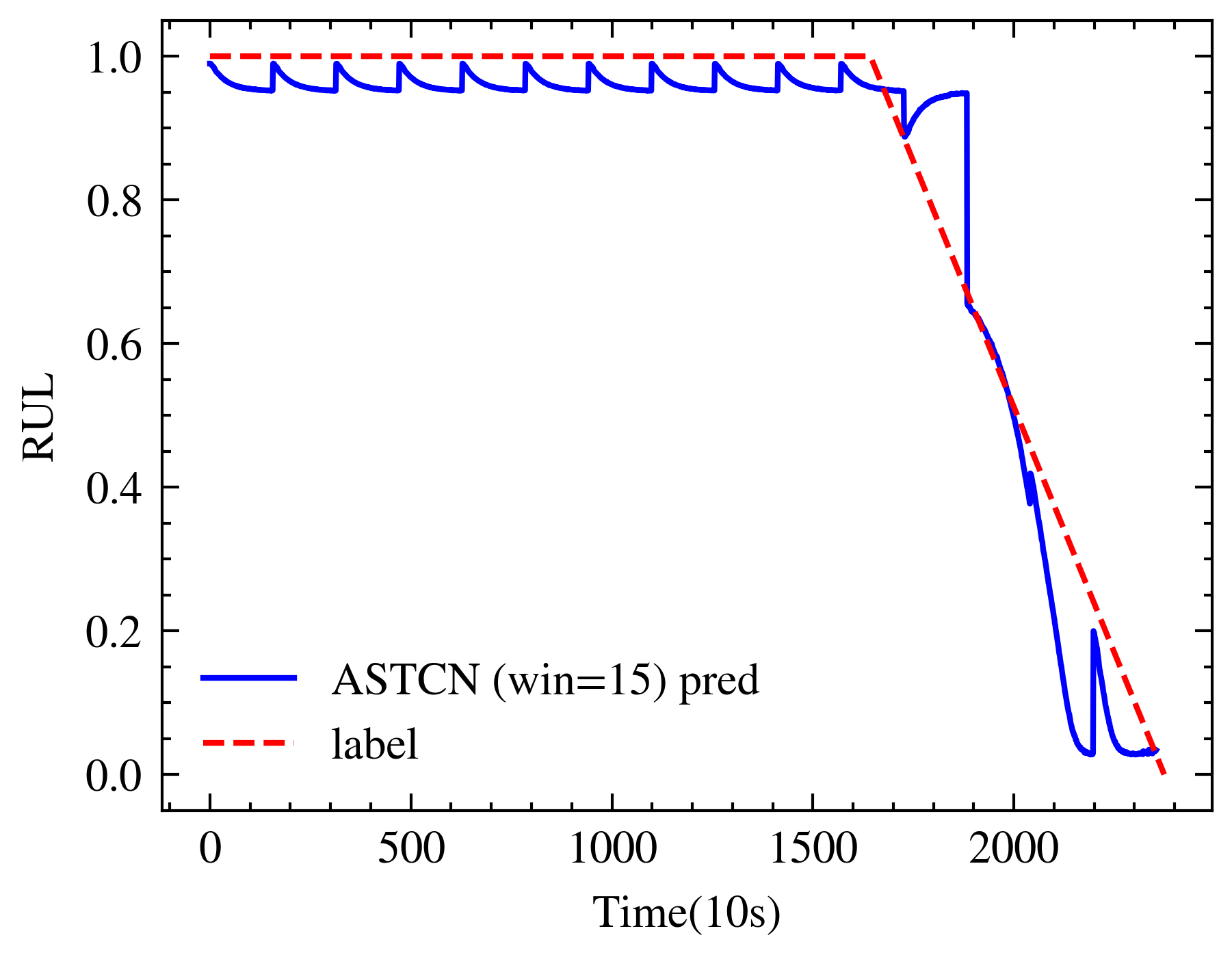} &
        \\
        (a)  & (b)  \\
    \end{tabular}
    \caption{Prediction results of different models in window size 15. (a) SKNET (b) ASTCN }
    \label{fig:pred15}

\end{figure*}

\begin{figure*}[htbp]
   \begin{tabular}{ccc}
        \includegraphics[width=0.5\linewidth]{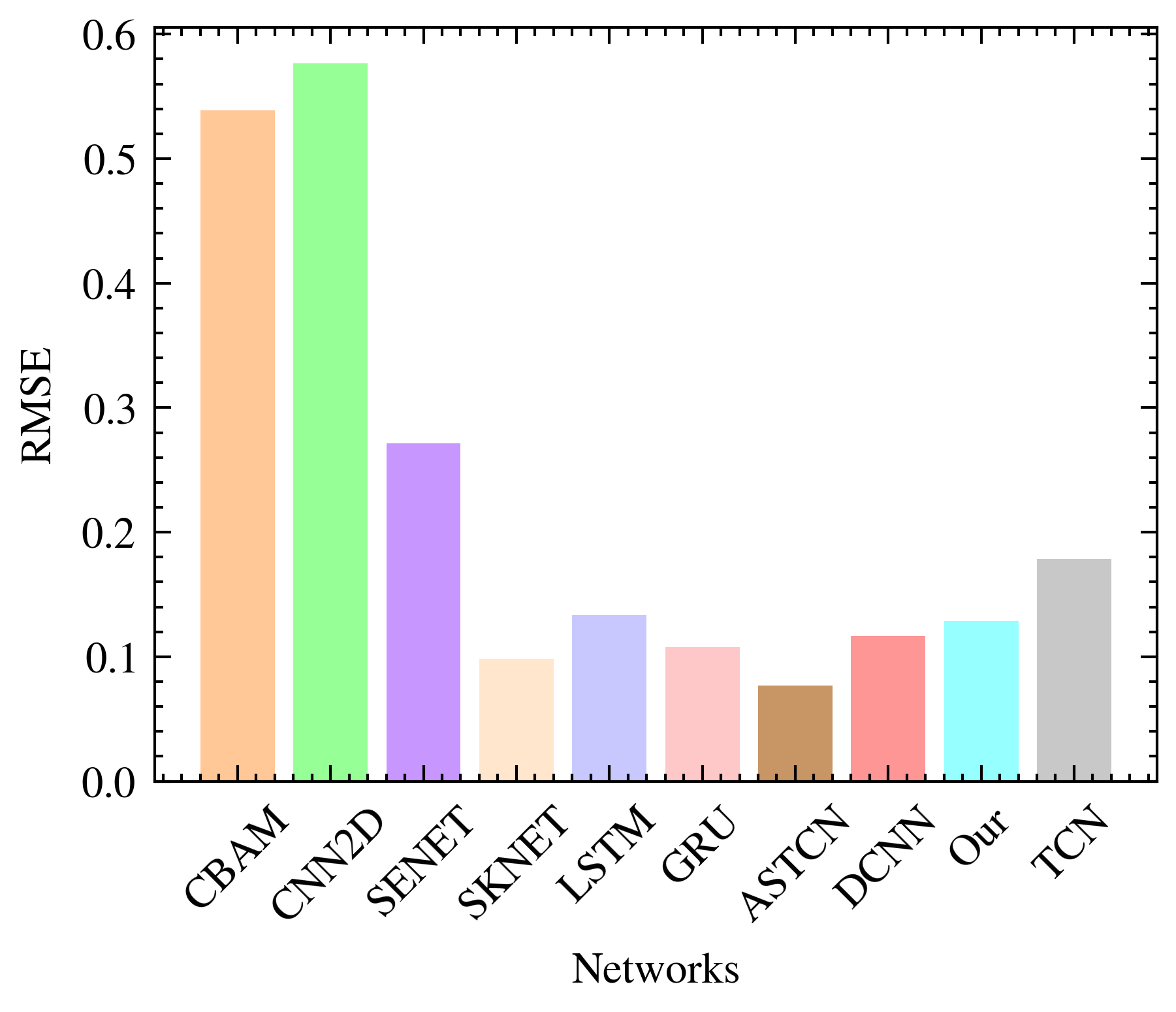} & \includegraphics[width=0.5\linewidth]{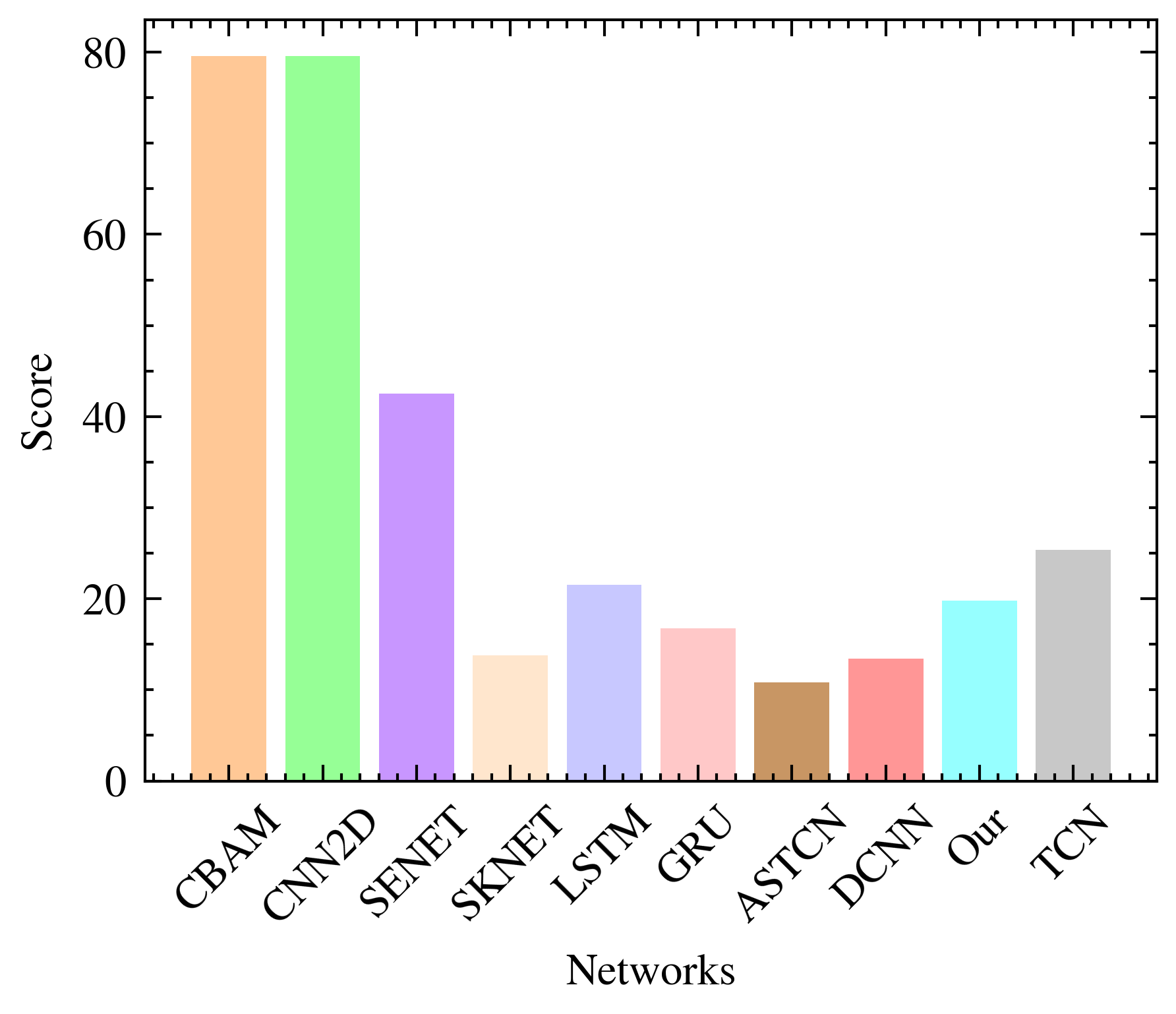} &
        \\
        (a)  & (b)  \\
    \end{tabular}
    \caption{Evaluation results of different models on bearing 1-1 under a window size of 15. (a) RMSE (b) Score }
    \label{fig:eval15}

\end{figure*}

Figure \ref{fig:eval15} compares the RMSE and Score differences of different models on bearings 1-3 under the same training conditions, with a window size of 15. We found that models with two-dimensional inputs consistently yield poorer results, with significantly higher RMSE and Score than other one-dimensional models. In fact, we tried different hyperparameters, but the two-dimensional models failed to generalize well. In this context, the numerical differences do not accurately reflect the models' capabilities. When the window size is small, the two-dimensional CNN models perform better, a conclusion that is the exact opposite for one-dimensional CNN models. Notably, at a window size of 15, the SKNet achieved a certain level of competitiveness. As shown in Figure \ref{fig:pred15}a, another advantage of the two-dimensional models is their smoother output, unlike the one-dimensional models, which exhibit a strong sense of paragraph segmentation. Among one-dimensional models, the ASTCN achieved the best results, also having the lowest RMSE and Score of all models. However, its overall performance is not outstanding, with significant prediction differences in both degradation and normal phases, as shown in Figure \ref{fig:pred15}b. The score of the DCNN is not much different from that of the SKNet, and other models have larger errors at this window size, making them not comparable. Due to the less-than-ideal performance of the models, we increased the window size.
\begin{figure*}[htbp]
   \begin{tabular}{ccc}
        \includegraphics[width=0.5\linewidth]{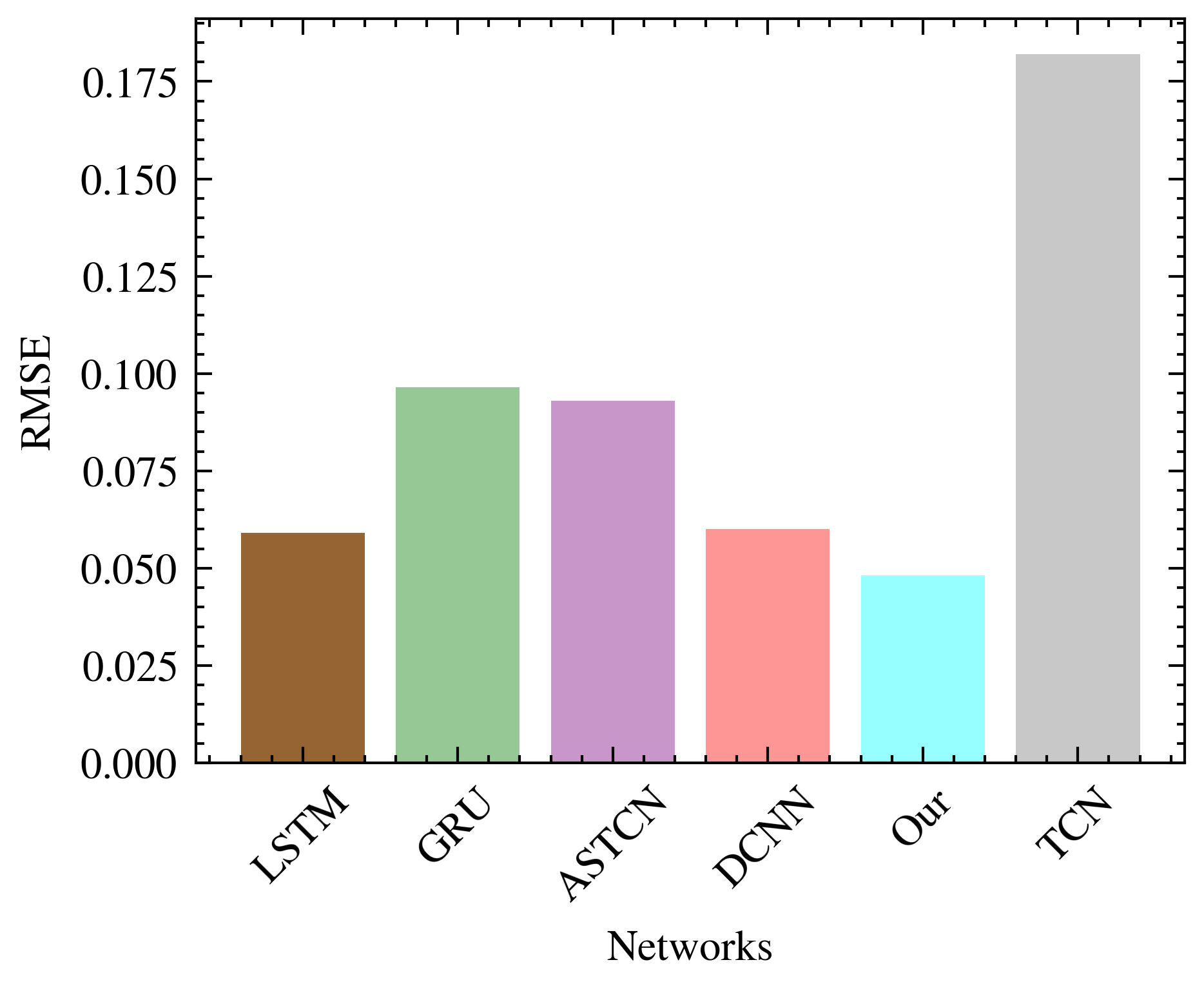} & \includegraphics[width=0.5\linewidth]{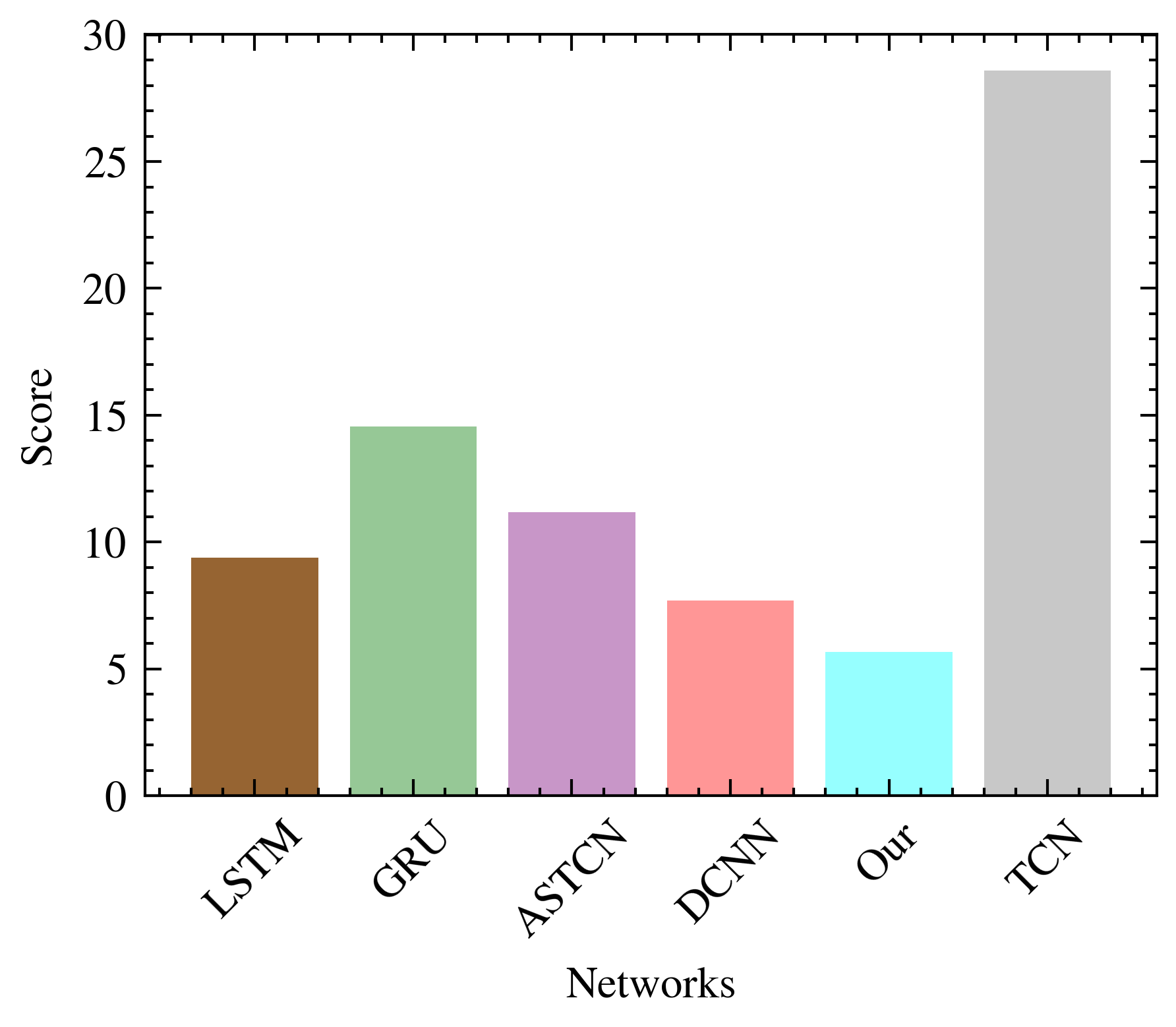} &
        \\
        (a)  & (b)  \\
    \end{tabular}
    \caption{Evaluation results of different models on bearing 1-1 under a window size of 45. (a) RMSE (b) Score }
    \label{fig:eval45}

\end{figure*}

\begin{figure*}[htbp]
   \begin{tabular}{ccc}
        \includegraphics[width=0.5\linewidth]{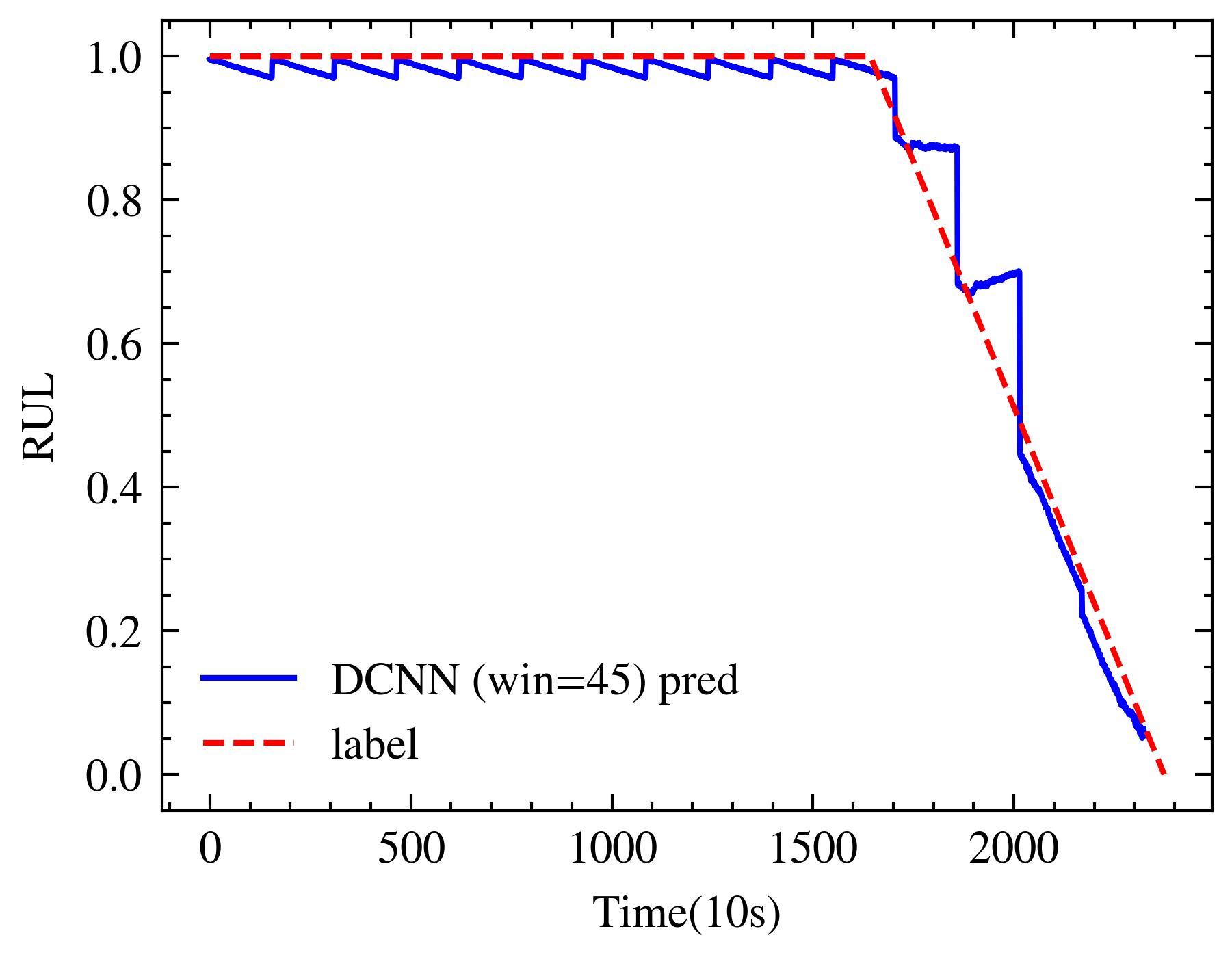} & \includegraphics[width=0.5\linewidth]{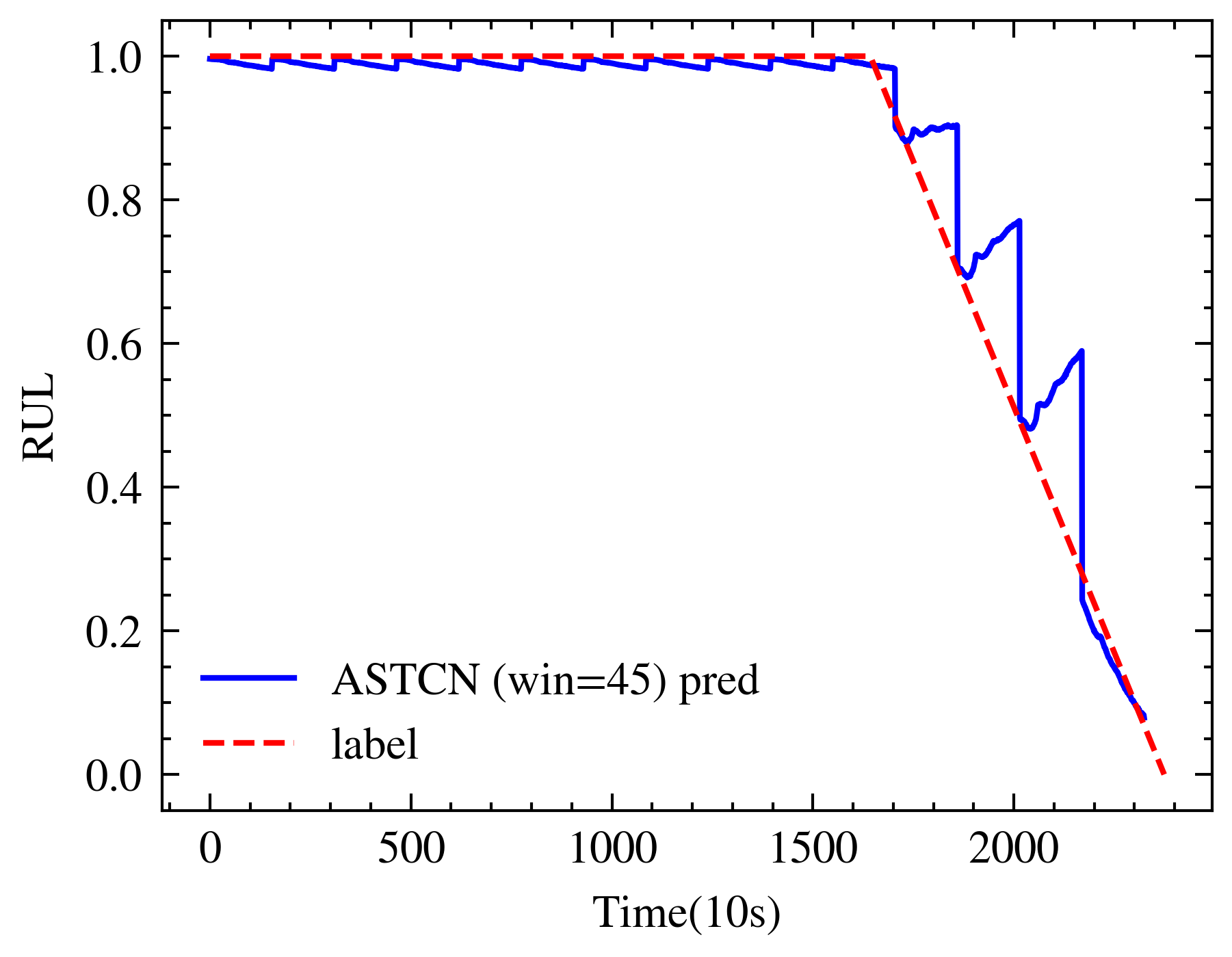} &
        \\
        (a)  & (b)  \\
    \end{tabular}
    \caption{Prediction results of different models in window size 45. (a) DCNN (b) ASTCN }
    \label{fig:pred45}

\end{figure*}

When the window size is set to 45, in the one-dimensional models, our model achieved the best results, with the lowest RMSE and Score. Following closely is the DCNN model, which has a very simple structure but impressive performance. Slightly more complex models like LSTM and GRU are always mediocre, failing to achieve the best results, but also not the worst. The score of ASTCN ranks fourth, but as observed from Figure \ref{fig:eval45}, ASTCN's prediction results are smoother than DCNN during the normal phase at HI=1, but fluctuate more intensely during the degradation phase, as shown in Figure \ref{fig:pred45}. The standard TCN model, however, still fails to predict accurately, and its performance at various window sizes is not ideal, indicating that this model needs further improvement. The two-dimensional models experienced memory overflow at this window size, hence there is no data available.

\subsection{Generalization test}
Well-trained models often demonstrate excellent generalization capabilities. Thus, in our paper, we chose not to fine-tune the model but instead directly applied it for predictions under operating condition 2. Specifically, bearings 2-1 to 2-4 were selected as the test set. After training on bearings 1-1 to 1-6, the models were tested directly without any further fine-tuning. Table \ref{tab:gen} presents the average test results of different backbone networks on bearings 2-1 to 2-4. It is evident that our proposed model achieved the lowest metrics in RMSE, MAE, and Score. DCNN and LSTM followed closely, with their RMSEs around 0.05, indicating a decent generalization capability. In contrast, the performance of two-dimensional networks was suboptimal, similar to their performance under operating condition 1. As for non-autoregressive neural networks, regardless of being one-dimensional or two-dimensional, they exhibited RMSEs above 0.3, suggesting minimal to no generalization ability. This is attributed to the lack of historical HI values in non-autoregressive settings, which are crucial for decision-making, coupled with significant differences between operating conditions. Overall, our multi-input autoregressive model demonstrated a significant gap in generalization capability compared to its non-autoregressive counterpart, and likewise, one-dimensional networks outperformed two-dimensional networks in this aspect.



\begin{table}[htbp]
\centering
\caption{Evaluation results of different models and different training methods. }
    \label{tab:gen}
\begin{tabularx}{\linewidth}{XXXX}
\toprule
    \textbf{Net}   & \textbf{MAE}   & \textbf{RMSE}  & \textbf{Score}    \\
\midrule
Our & \textbf{0.039} & \textbf{0.049} & \textbf{3.82}    \\
CBAM       & 0.262 & 0.348 & 34.80 \\
SKNET      & 0.258 & 0.355 & 32.52 \\
SENET      & 0.127 & 0.161 & 15.53 \\
CNN        & 0.228 & 0.287 & 29.53 \\
LSTM       & 0.040 & 0.050 & 4.14  \\
GRU        & 0.057 & 0.069 & 6.00  \\
ASTCN      & 0.064 & 0.083 & 6.64  \\
DCNN       & 0.039 & 0.051 & 3.90  \\
TCN        & 0.127 & 0.153 & 14.45 \\
CBAM(NAR)  & 0.324 & 0.446 & 38.77 \\
SKNET(NAR) & 0.342 & 0.467 & 40.75 \\
SENET(NAR) & 0.354 & 0.482 & 42.62 \\
LSTM(NAR)  & 0.409 & 0.478 & 34.27 \\
GRU(NAR)   & 0.294 & 0.332 & 29.00 \\
ASTCN(NAR) & 0.328 & 0.418 & 39.47 \\
\bottomrule
\end{tabularx}
\end{table}

\section{Conclusion}
This paper introduces a novel multi-input autoregressive model designed to address the challenge of predicting the remaining useful life (RUL) of bearings. Our method not only uses vibration signals as inputs but also incorporates previously predicted health indicator (HI) values. By employing feature fusion, the model outputs HI values for the current window. A key innovation of our approach is the use of a segmentation method in conjunction with multiple training iterations to overcome the problem of error accumulation in autoregressive models. The segmentation method involves dividing a complete bearing dataset into multiple independent samples, each undergoing individual training. Furthermore, for each segment, we select a subset of samples and divide them into parts for multiple training iterations. In these iterations, only the model is updated, not the label values, until a predetermined number of training cycles is completed.
Empirical evaluation on the PMH2012 dataset demonstrates that our model, using the same autoregressive method, consistently achieved the lowest RMSE and Score metrics compared to other backbone networks. Notably, when contrasted with autoregressive models that use label values as inputs and traditional non-autoregressive networks, our model exhibited superior generalization ability, significantly leading in RMSE and Score metrics.

In future work, we aim to explore more sophisticated backbone network architectures to further enhance the predictive performance and training stability of our network. Additionally, we plan to experiment with alternative methods of generating HI curves to more accurately guide the network.



\section*{Acknowledgments}
The research was partially supported by the Key Project of Natural Science Foundation of China (No. 61933013), the Science and Technology Innovation Strategy Project of Guangdong Provincial(No. 2023S002028) and the special projects in key fields of ordinary universities in Guangdong Province(No.2023ZDZX3015). We also appreciate the technical support of the Guangdong Provincial Key Laboratory of Petrochemical Equipment and Fault Diagnosis.



\bibliographystyle{IEEEtran}
\bibliography{IEEEabrv,paper}

\end{document}